\documentclass[runningheads]{llncs}

\usepackage{eccv}

\usepackage{eccvabbrv}

\usepackage{graphicx}
\usepackage{booktabs}

\usepackage[accsupp]{axessibility}  % Improves PDF readability for those with disabilities.

\usepackage{hyperref}

\usepackage{orcidlink}

\usepackage{times} % assumes new font selection scheme installed
\usepackage{amsmath} % assumes amsmath package installed
\usepackage{amssymb}  % assumes amsmath package installed
\usepackage{booktabs}
\usepackage[ruled,linesnumbered,vlined]{algorithm2e}
\usepackage{caption}
\usepackage{pifont}
\usepackage{setspace}
\usepackage{multirow}
\usepackage{diagbox}
\usepackage{xcolor}
\usepackage{booktabs}
\usepackage{colortbl}
\usepackage{adjustbox}
\definecolor{modelAHdr}{HTML}{C8DDFB}   % OpenVLA - 浅蓝
\definecolor{modelBHdr}{HTML}{CBCEF0}   % OpenVLA-OFT - 浅靛蓝
\definecolor{modelCHdr}{HTML}{D5CCF0}   % UniVLA - 浅紫
\definecolor{modelDHdr}{HTML}{E0D4F5}   % pi0.5 - 浅violet

\newcommand{\dd}[1]{{\scriptsize\,($\uparrow$\!#1)}}     % delta up (FR increased)           
\newcommand{\du}[1]{{\scriptsize\,($\downarrow$\!#1)}}   % delta down (FR decreased, rare)    

\usepackage[dvipsnames]{xcolor}

\begin{document}

\title{Eva-VLA: Evaluating Vision-Language-Action Models' Robustness Under Real-World Physical Variations}

% \author{
% Hanqing Liu$^{1,2,3}$, Shouwei Ruan$^{4}$, Jiahuan Long$^{1,2,3}$,  Junqi Wu$^{1,2,3}$, Jiacheng Hou$^{2,3}$, Huili Tang$^{1,2,3}$, Tingsong Jiang$^{2,3*}$, Weien Zhou$^{2,3}$, Wen Yao$^{2,3*}$ \\
% $^{1}$MoE Key Lab of Artificial Intelligence, Al Institute, Shanghai Jiao Tong University \\
% $^{2}$Defense Innovation Institute, Chinese Academy of Military Science\\
% $^{3}$Intelligent Game and Decision Laboratory\\
% $^{4}$Institute of Artificial Intelligence, Beihang University\\
% \scriptsize{\texttt{hanqingliu@sjtu.edu.cn,tingsong@pku.edu.cn,wendy0782@126.com}}
% }

% --- 作者部分 ---
\author{Hanqing Liu\inst{1,2,3} \and
Shouwei Ruan\inst{4} \and
Jiahuan Long\inst{1,2,3} \and
Junqi Wu\inst{1,2,3} \and
Jiacheng Hou\inst{2,3} \and
Huili Tang\inst{1,2,3} \and
Tingsong Jiang\inst{2,3} \and
Weien Zhou\inst{2,3} \and
Wen Yao\inst{2,3}}

% --- 运行页作者简写 ---
\authorrunning{H.~Liu et al.}

% --- 机构与邮箱部分 ---
\institute{
MoE Key Lab of Artificial Intelligence, AI Institute, Shanghai Jiao Tong University \and
Defense Innovation Institute, Chinese Academy of Military Science \and
Intelligent Game and Decision Laboratory \and
Institute of Artificial Intelligence, Beihang University\\
\email{hanqingliu@sjtu.edu.cn, tingsong@pku.edu.cn, wendy0782@126.com}
}


% Second Author\inst{2,3}\orcidlink{1111-2222-3333-4444} \and
% Third Author\inst{3}\orcidlink{2222--3333-4444-5555}}


% Springer Heidelberg, Tiergartenstr.~17, 69121 Heidelberg, Germany

\maketitle

\begin{abstract}

Vision-Language-Action (VLA) models have emerged as promising solutions for robotic manipulation, yet their robustness to real-world physical variations remains critically underexplored. To bridge this gap, we propose \textbf{Eva-VLA}, the first unified framework to systematically evaluate the robustness of VLA models by formulating uncontrollable physical variations as continuous optimization problems. Specifically, our framework addresses two fundamental challenges in VLA models' physical robustness evaluation: \textbf{\textit{1)}} how to systematically characterize diverse physical perturbations encountered in real-world deployment while maintaining reproducibility, and \textbf{\textit{2)}} how to efficiently discover worst-case scenarios without incurring prohibitive real-world data collection costs. \textbf{\textit{To tackle the first challenge}}, we decouple real-world variations into three key dimensions: 3D object transformations that affect spatial reasoning, illumination changes that challenge visual perception, and adversarial regions that disrupt scene understanding. \textbf{\textit{For the second challenge}}, we introduce a continuous black-box optimization mechanism that maps these perturbations into a continuous parameter space, enabling the systematic exploration of worst-case scenarios. Extensive experiments validate the effectiveness of our approach. Notably, OpenVLA exhibits an average failure rate of over 90\% across three physical variations on the LIBERO-Long task, exposing critical systemic fragilities. Furthermore, applying the generated worst-case scenarios during adversarial training quantifiably increases model robustness, validating the effectiveness of this approach. Our evaluation exposes the gap between laboratory and real-world conditions, while the Eva-VLA framework can serve as an effective data augmentation method to enhance the resilience of robotic manipulation systems.
\keywords{Vision-Language-Action Model \and Adversarial Robustness}
\end{abstract}

\section{Introduction}
Vision-Language-Action (VLA) models represent a paradigm shift in robotic manipulation, integrating visual perception, language understanding, and action generation into unified end-to-end systems~\cite{ma2024survey,brohan2023rt}. Recent deployments across manufacturing~\cite{li2024cogact,bharadhwaj2024roboagent,qin2023anyteleop}, healthcare~\cite{li2024robonurse,fu2024mobile}, and service robotics~\cite{kim2024openvla,black2410pi0} demonstrate their transformative potential. However, in real-world deployments, VLA models inevitably face challenging physical variations, such as spatial transformations, illumination variations, and visual disruptions, which can dramatically alter robot behavior without being immediately detectable, posing significant safety risks. Therefore, it is crucial to investigate VLA robustness across various physical conditions systematically.

Existing research has explored the robustness of VLA-based robotic systems through approaches like adversarial patches~\cite{wang2024exploring}, which generate localized perturbations via gradient-based white-box attacks to achieve visual interference. However, these methods suffer from critical limitations: they violate physical plausibility constraints and fail to capture the rich spectrum of real-world physical variations. Moreover, their reliance on gradient access restricts their applicability to black-box deployment scenarios. To overcome these limitations, we aim to develop a model-agnostic framework that generates diverse and physically realistic variations for comprehensively evaluating VLA robustness under black-box settings. Achieving this, however, requires addressing two key challenges: \textbf{(1) \textit{How to systematically characterize diverse physical variations encountered in real-world deployments while maintaining evaluation reproducibility?}} \textbf{(2) \textit{How to efficiently discover worst-case scenarios without incurring prohibitive real-world data collection costs?}}

\begin{figure}[t!]
    \centering
    \includegraphics[width=0.95\linewidth]{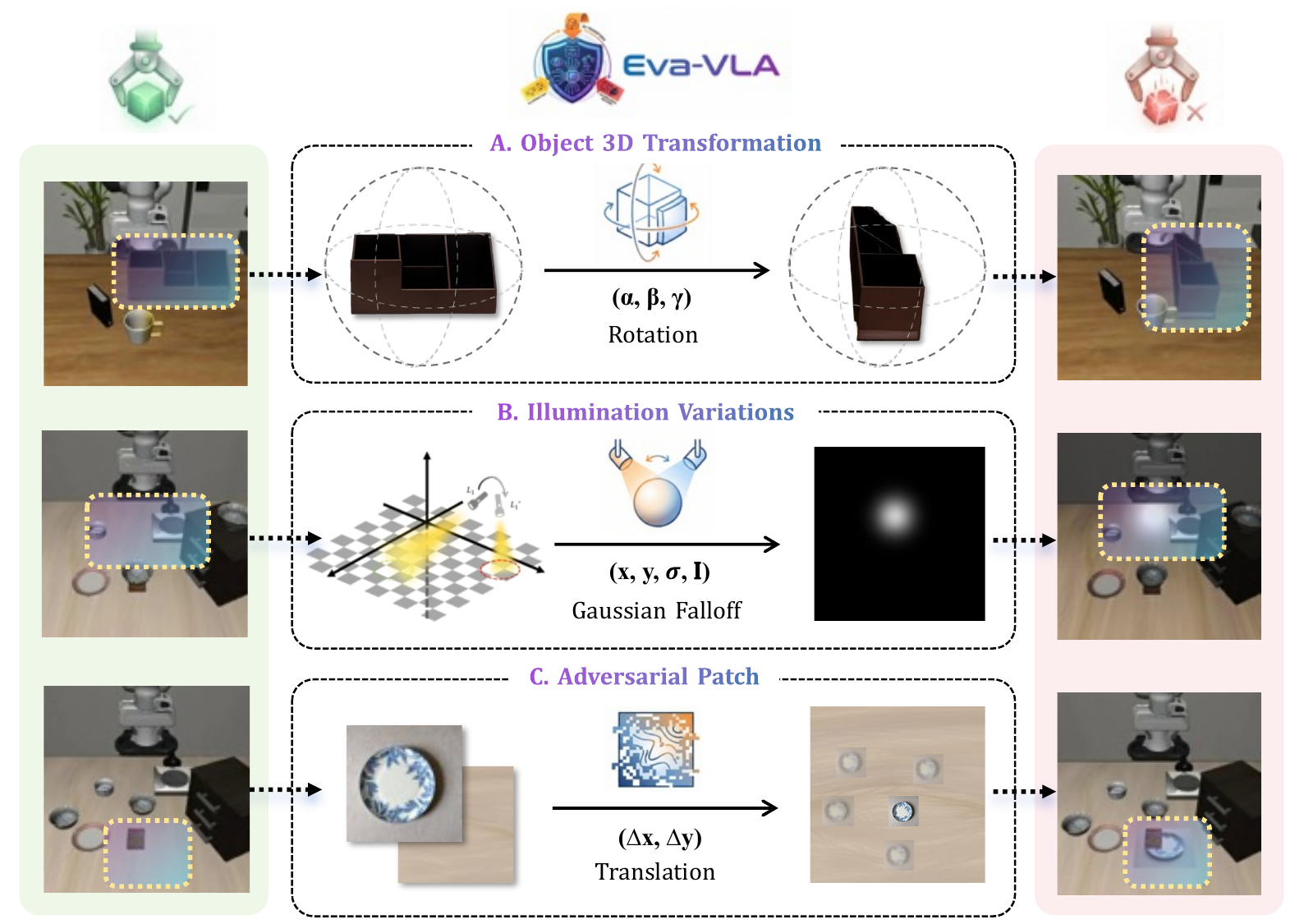}
    \caption{\textbf{Visualization of three categories of physical variations.} Object 3D transformations through rotation parameters ($\alpha$, $\beta$, $\gamma$) that alter object 3D poses in the scene \textbf{(Top)}. Illumination variations modeled as Gaussian falloff functions with parameters ($x$, $y$, $\sigma$, $I$) controlling illumination position, radius, and intensity \textbf{(Middle)}. Adversarial patches with translation parameters ($\Delta x$, $\Delta y$) that introduce visual disruptions at critical locations in the scene \textbf{Bottom)}.}
    \label{fig:display}
\end{figure}

To address these challenges, we propose \textbf{\textit{Eva-VLA}}, a unified framework for evaluating vision-language-action models' robustness. Our key innovation is the design of a physics-aware, gradient-free vulnerability discovery pipeline that systematically maps unstructured physical variations into a tractable, continuous search space. First, as shown in Fig.~\ref{fig:display}, we decompose real-world variations into three distinct domains: object 3D transformations parameterized with rotation angles ($\alpha$, $\beta$, $\gamma$), illumination variations defined by point light parameters including position ($x$, $y$), radius ($\sigma$), and intensity ($I$), and adversarial patch placement specified by spatial offsets ($\Delta x$, $\Delta y$). This parameterization enables systematic exploration of the variation space while maintaining physical plausibility through explicit constraints. Second, to overcome the black-box nature of VLA models and non-differentiable simulation environments, we employ Covariance Matrix Adaptation Evolution Strategy (CMA-ES)~\cite{hansen2016cma}, a gradient-free optimization algorithm, to efficiently discover worst-case scenarios by iteratively optimizing physical variation parameters. This approach enables comprehensive vulnerability assessment without requiring model gradients or expensive real-world data collection. Furthermore, to validate the practical utility of the discovered worst-case scenarios, we employ them as adversarial examples to enhance model robustness. This supplementary adversarial training serves as an empirical verification, demonstrating that the vulnerabilities identified by our framework are not merely theoretical flaws, but high-quality, actionable data that can be utilized to improve model robustness.

Our main contributions are summarized as follows:

\begin{enumerate}
    \item We systematically categorize complex physical variations into three distinct dimensions: object 3D transformations, illumination, and adversarial patches, providing a comprehensive framework for evaluating physical robustness.
    \item We propose \textbf{\textit{Eva-VLA}}, a physics-aware, gradient-free framework that formulates uncontrollable physical variations as a continuous optimization problem. By leveraging a reproducible simulation environment, it enables the efficient discovery of worst-case scenarios while bypassing prohibitive real-world costs.
    \item Extensive evaluations on the LIBERO~\cite{liu2023libero} benchmark expose severe vulnerabilities across leading VLA models (e.g., OpenVLA~\cite{kim2024openvla}, UniVLA~\cite{bu2025univla}, $\pi_{0.5}$~\cite{intelligence2025pi_}). Crucially, enhencing models with our generated variations significantly improves robustness, validating the practical necessity of our framework.
\end{enumerate}

\begin{figure*}[t]
  \centering
  \includegraphics[width=\textwidth]{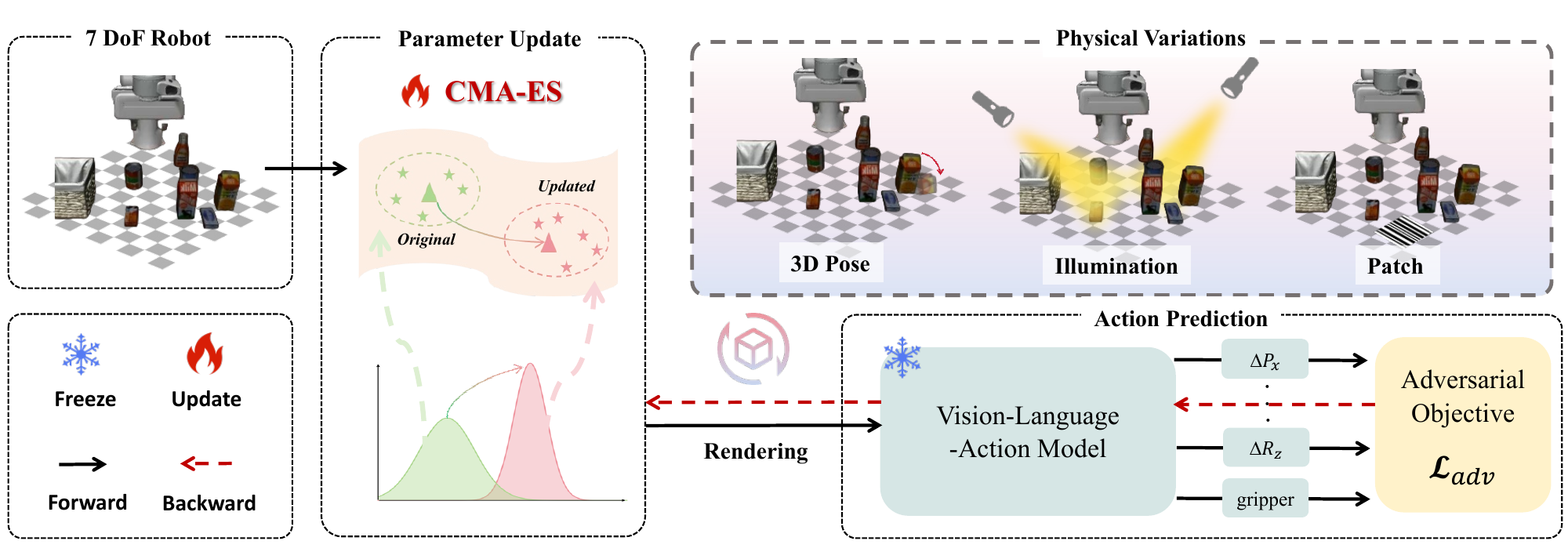}
  % \vspace{-2ex}
  \caption{\textbf{Overview of the Proposed Eva-VLA Framework.} To capture worst-case physical variations, discrete transformations in three critical domains are parameterized and their distributions optimized through a query-based method, maximizing prediction errors of vision-language-action models to reduce task success rates.}
  % \vspace{-3ex}
    \label{fig:framework}
\end{figure*}

\section{Related Work}
\subsection{Vision-Language-Action Models}
Recent studies \cite{li2023vision, brohan2023rt, kim2024openvla, li2024cogact, zheng2024tracevla, li2024towards, zheng2025universal, pertsch2025fast} have explored extending pre-trained Vision-Language Models (VLMs) to generate robot actions, thereby contributing to the development of generalist robotic policies. RT-2 \cite{brohan2023rt} pioneered this approach by fine-tuning PaLI-X \cite{chen2023pali} with robot actions discretized into 256 bins as tokens. Building on this foundation, OpenVLA \cite{kim2024openvla} applies a similar discretization but fine-tunes Prismatic VLM \cite{karamcheti2024prismatic} specifically on the OXE dataset \cite{vuong2023open}, which contains data from 22 robot embodiments. The OpenVLA architecture has spawned several improvements. For example, CogACT \cite{li2024cogact} incorporates a diffusion-based action module for better continuous action modeling. TraceVLA \cite{zheng2024tracevla} introduces visual trace prompting for enhanced spatial-temporal awareness, achieving 3.5x performance gains on real robot tasks. SpatialVLA \cite{qu2025spatialvla} adds 3D position encoding and adaptive action grids for improved spatial understanding. Notably, OpenVLA-OFT \cite{kim2025fine} optimizes fine-tuning through parallel decoding and action chunking, achieving 26x speedup. Beyond the OpenVLA lineage, the $\pi$ series has emerged as a powerful alternative by utilizing flow matching for continuous action generation. Its advanced iteration, $\pi_{0.5}$ \cite{intelligence2025pi_}, significantly enhances open-world generalization through a hierarchical architecture that couples high-level semantic reasoning with rapid, low-level motor control. Concurrently, UniVLA \cite{bu2025univla} tackles cross-embodiment generalization by deriving task-centric latent actions directly from internet-scale videos without requiring action labels. By establishing a latent action model within the DINO feature space and incorporating language instructions, UniVLA effectively decouples task-relevant dynamics from extraneous visual changes, enabling highly efficient policy learning and downstream deployment. To ensure a comprehensive and rigorous evaluation of our framework, we select the currently most advanced state-of-the-art models, including the OpenVLA series~\cite{kim2024openvla}~\cite{kim2025fine}, $\pi_{0.5}$~\cite{intelligence2025pi_}, and UniVLA~\cite{bu2025univla}, as our target models. This selection covers a diverse spectrum of cutting-edge mechanisms (ranging from standard discrete tokenization and continuous flow-matching to unsupervised latent action learning from cross-embodiment videos) while benefiting from widespread open-source community adoption, guaranteeing highly representative and reproducible comparisons.

\subsection{Adversarial Attacks in Robotic Systems}

Adversarial attacks play a crucial role in assessing the vulnerabilities of machine learning models, particularly in robotics, where models operate in dynamic, real-world environments. Traditional gradient-based, pixel-level attacks \cite{du2022physical,goodfellow2014explaining,liu2019perceptual,madry2017towards,su2019one} exploit model gradients to calculate malicious perturbations, achieving high success rates in controlled digital settings. However, when applied to real-world scenarios, these attacks often face significant challenges due to the complexity and variability of physical environments. For physical-world attacks, patch-based methods \cite{athalye2018synthesizing,brown2017adversarial} have emerged as a practical solution, with recent works examining illumination variations~\cite{liu2025lighting,chen2025robotwin} and viewpoint variations~\cite{ruan2024advdreamer} impact model robustness. Since VLA models directly output control signals for robotic actions through end-to-end learning, attacks targeting the visual input can significantly affect the final action outputs. In this context, Cheng et al. \cite{cheng2024manipulation} introduced the Physical Vulnerability Evaluating Pipeline (PVEP) to systematically assess VLA robustness against physical visual threats. However, while PVEP provides a foundational analysis of visual threats in robotic manipulation, it predominantly relies on predefined threat categories and constrained 2D perturbations, which may not fully capture the continuous, multi-dimensional nature of real-world physical shifts. Wang et al. \cite{wang2024exploring} explored the vulnerabilities of VLA systems by placing a colored patch within the camera’s field of view. While this approach led to a decrease in task success rates, it relied on white-box optimization and produced patches that were easily detectable by the human eye. To bridge these critical gaps, we propose \textbf{\textit{Eva-VLA}}, the first comprehensive framework to systematically evaluate VLA robustness against continuous, multi-dimensional physical transformations.

% By optimizing 3D object poses, illumination variations, and adversarial patches without relying on white-box access, our approach practically addresses the complexities of real-world deployment challenges.

\section{Methodology}

In this section, we first provide an overview of VLA models and their underlying principles. Then, we present our Eva-VLA framework (illustrated in Fig.~\ref{fig:framework}), detailing the parameterization of three categories of physical variations and the formulation of our adversarial objectives. Finally, we describe the optimization algorithm that efficiently discovers adversarial distributions over these physical variations.

\subsection{Preliminaries}

Vision-Language-Action (VLA) models are
built on large language models integrated with visual encoders, enabling end-to-end task execution by jointly processing visual perception, linguistic instructions, and action generation. Formally, a VLA model can be defined as: $
F: \mathcal{V} \times \mathcal{L} \rightarrow \mathcal{A}
$, 
where $\mathcal{V}$ represents the visual observation space, $\mathcal{L}$ denotes the language instruction space, and $\mathcal{A}$ is the action output space. In this work, we focus on a 7 degrees of freedom (DoF) robotic manipulator~\cite{haddadin2022franka}, where the action space encodes both end-effector motion and gripper control. The output action vector can be specified as:
\begin{equation}
A = [\Delta P_x, \Delta P_y, \Delta P_z, \Delta R_x, \Delta R_y, \Delta R_z, gripper],
\end{equation}
where $\Delta P_{x,y,z} \in \mathbb{R}^3$ and $\Delta R_{x,y,z} \in \mathbb{R}^3$ represent the positional and rotational changes along the Cartesian axes, and $gripper \in \mathbb{R}$ controls the gripper states~\cite{kim2024openvla}.

\subsection{Eva-VLA Framework}
To comprehensively evaluate VLA models' robustness, we introduce Eva-VLA, a unified framework (illustrated in Fig.~\ref{fig:framework}) that parameterizes three distinct categories of physical variations. These categories span critical aspects of the environmental shifts commonly encountered in real-world robotic deployments. \ding{182} \textbf{Parametrization of 3D Transformation.} We focus on rigid 3D transformations of important objects in the scene, such as the rotation, since it reflects the most typical 3D changes observed in real-world environments. Formally, we define a 3-dimensional vector $\boldsymbol{\Theta}=\{\alpha,\beta,\gamma\}$ to uniquely parameterize any arbitrary rotation, representing the Tait-Bryan angles (yaw, pitch, roll) in the $Z$-$Y$-$X$ sequence. In order to comply with the physical laws of the space in which the target is placed, we constrain this transformation to a bounded interval $[\boldsymbol{\Theta}_{\min}, \boldsymbol{\Theta}_{\max}]$. \ding{183} \textbf{Parametrization of Illumination Variations.} In this section, we primarily focus on the illumination variations in real-world environments. Formally, we define a 4-dimensional vector $\boldsymbol{\Lambda}=\{x,y,\sigma,I\}$ to parameterize the illumination environment. Then, to simulate realistic light behavior, we apply a Gaussian falloff function to model the spatial distribution of the light source, which can be represented as:
\begin{equation}
L(z) = I \cdot \exp\left(-\frac{\| z - (x, y) \|^2}{2\sigma^2}\right),
\label{problem2}
\end{equation}
where $z$ represents any point in the scene. The parameters $(x, y)$, $\sigma$, and $I$ from $\boldsymbol{\Lambda}$ directly control the center coordinates, spread radius, and intensity of the light source, respectively. \ding{184} \textbf{Parametrization of Adversarial Patch.} Rather than optimizing patch textures~\cite{wang2024exploring}, we employ natural images (e.g., barcodes, QR codes, everyday images) and optimize their spatial placement on the tabletop surface. This approach ensures physical realizability through standard UV mapping while maximizing perceptual impact, as the table occupies a significant portion of the robot's visual field during manipulation. Formally, we define a 2-dimensional vector: $\boldsymbol{\phi} = \{x, y\} \in [\boldsymbol{\phi}_{min},\boldsymbol{\phi}_{max}]$, where $(x, y)$ denotes the position of the adversarial patch in the table texture. This constraint ensures patches remain within the robot's primary workspace, reducing optimization queries while maintaining adversarial effectiveness. 

Next, we formalize the adversarial objective to systematically quantify the degradation of VLA model performance under these physical variations. To provide a continuous guiding signal for the optimization algorithm, we utilize cosine similarity to measure the deviation of the predicted action vectors from the nominal, clean trajectory. This dense metric effectively steers the search process away from the model's standard execution behavior. However, robotic manipulation is inherently characterized by policy redundancy, meaning multiple distinct action sequences can successfully accomplish the same task. Consequently, a pronounced divergence in the action space does not definitively signify a task failure; the model may simply be executing an alternative valid trajectory. To account for this phenomenon and ensure the optimization specifically targets genuine execution breakdowns, we augment the objective with a substantial terminal reward. This reward is heavily weighted and triggered exclusively when the physical perturbation results in an ultimate task failure. Therefore, our complete adversarial objective $\mathcal{L}_{adv}$ is formulated as:
\begin{equation}
\mathcal{L}_{adv} = -\sum_{i=1}^{N} \cos(A^i_{clean}, A^i_{adv}) + \lambda \cdot \mathbb{I}_{fail},
\end{equation}
where $N$ denotes the total number of evaluation timesteps, and $A^i_{adv} = F(T(X^i, \boldsymbol{\mathcal{T}}), Y)$ represents the $i$-th adversarial action vector generated under the parameterized physical variations $\boldsymbol{\mathcal{T}} \in \{\boldsymbol{\Theta}, \boldsymbol{\Lambda}, \boldsymbol{\phi}\}$. The term $\mathbb{I}_{fail} \in \{0, 1\}$ is a binary indicator function that equals $1$ if the robotic task ultimately fails, and $\lambda$ is a large positive scalar hyperparameter ensuring that actual task failures yield the maximum optimization reward.Thus, the objective of Eva-VLA is to discover an optimal distribution $p^*(\boldsymbol{\mathcal{T}})$ over the transformation parameters that maximizes this expected adversarial impact:

\begin{equation}
p^*(\boldsymbol{\mathcal{T}}) = \arg\max_{p(\boldsymbol{\mathcal{T}})} \mathbb{E}_{\boldsymbol{\mathcal{T}} \sim p(\boldsymbol{\mathcal{T}})} \left[ \mathcal{L}_{adv}(A_{clean}, A_{adv}) \right].
\end{equation}

% Then, we formulate the adversarial objective to systematically evaluate how these variations impact VLA model behavior. Specifically, our adversarial objective $\mathcal{L}_{adv}$ is to cause the 7-Dof robot arm to output incorrect action prediction vectors, which can be expressed as:
% \begin{equation}
% \mathcal{L}_{adv} = -\sum_{i=1}^Ncos(A^i_{clean}, A^i_{adv}),
% \end{equation}
% where \(N\) denotes the total number of action vector, \(A^i\) denotes the i-th action vector, which can be represented as: 
% \begin{equation}
% A_{adv}^i = F(T(X^i,\boldsymbol{\mathcal{T}}),Y),
% \end{equation}
% where  \(F\) represents the specified VLA models, $T(\cdot)$ is a transformation function, \(X\) denotes the input image, \(Y\) denotes the input instruction. Thus, the objective of Eva-VLA is to discover an optimal distribution $p^*(\boldsymbol{\mathcal{T}})$ over transformation parameters $\boldsymbol{\mathcal{T}} \in \{\boldsymbol{\Theta}, \boldsymbol{\Lambda}, \boldsymbol{\phi}\}$. The optimization seeks to ensure that parameters sampled from this distribution maximize adversarial impact on VLA models. This can be formulated as:
% \begin{equation}
% p^*(\boldsymbol{\mathcal{T}}) = \arg\max_{p(\boldsymbol{\mathcal{T}})}\mathbb{E}_{\boldsymbol{\mathcal{T}}\sim p(\boldsymbol{\mathcal{T})}}[\mathcal{L}_{adv}(A_{clean},A_{adv})].
% \end{equation}

\subsection{Query-Based Optimizing Algorithm}

In this section, we present our query-based optimization algorithm designed to identify the most potent physical variations, denoted as $\boldsymbol{\mathcal{T}}^*$. Rather than relying on point-estimate optimization to find a single worst-case configuration, we frame this as a distribution search problem. This probabilistic approach offers two key advantages: \textbf{1)} It enables diverse exploration of the configuration space, helping identify regions where optimal adversarial conditions exist; \textbf{2)} It facilitates faster convergence to regions containing effective configurations while reducing the risk of local optima.

Building a unified search framework across the diverse physical domains—3D transformations, illumination variations, and adversarial patch—requires modeling the search process probabilistically. This parameterization transforms the discrete search problem into a continuous optimization problem, enabling efficient exploration of the adversarial space. Specifically, each variation type $\boldsymbol{\mathcal{T}} \in \{\boldsymbol{\Theta}, \boldsymbol{\Lambda}, \boldsymbol{\phi}\}$ is modeled as a multivariate Gaussian distribution:
$\boldsymbol{\mathcal{T}} \sim \mathcal{N}(\boldsymbol{\mu}_{\boldsymbol{\mathcal{T}}}, \boldsymbol{\Sigma}_{\boldsymbol{\mathcal{T}}}^2 \boldsymbol{C}_{\boldsymbol{\mathcal{T}}})$, 
where $\boldsymbol{\mu}_{\mathcal{T}}$ represents the mean configuration, $\boldsymbol{\Sigma}_{\mathcal{T}}$ controls the exploration scale, and $\boldsymbol{C}_{\mathcal{T}}$ captures parameter correlations. While all three variation types share the same initial distribution structure, they evolve independently during optimization to discover variation-specific adversarial patterns.

The iterative refinement of these distribution parameters is driven by the Covariance Matrix Adaptation Evolution Strategy (CMA-ES)~\cite{hansen2016cma, golovin2017google}. This gradient-free evolutionary approach is intrinsically suited to our black-box evaluation paradigm. Relying solely on model inference outputs, CMA-ES completely bypasses the computational and practical hurdles of requiring differentiable simulation environments or white-box access to the underlying gradients of foundation VLA models.

Furthermore, the optimization loop integrates two key techniques to maximize search efficiency: \ding{182} \textbf{Learning Rate Adaptation (LRA):} This technique dynamically adjusts the step size $\boldsymbol{\Sigma}$ during optimization to balance exploration and exploitation, thereby improving convergence speed. \ding{183} \textbf{Early Stopping Policy:} Monitors the convergence criterion and terminates optimization when improvements plateau, preventing computational waste on marginal gains. The optimization process iterates through three main steps: sampling candidate configurations from the current distribution, evaluating adversarial effectiveness through VLA model queries, and updating distribution parameters based on the most successful candidates. We defer the formal algorithm and comprehensive procedural details to Appendix A.

\begin{figure*}[t]
  \centering
  \includegraphics[width=0.9\textwidth]{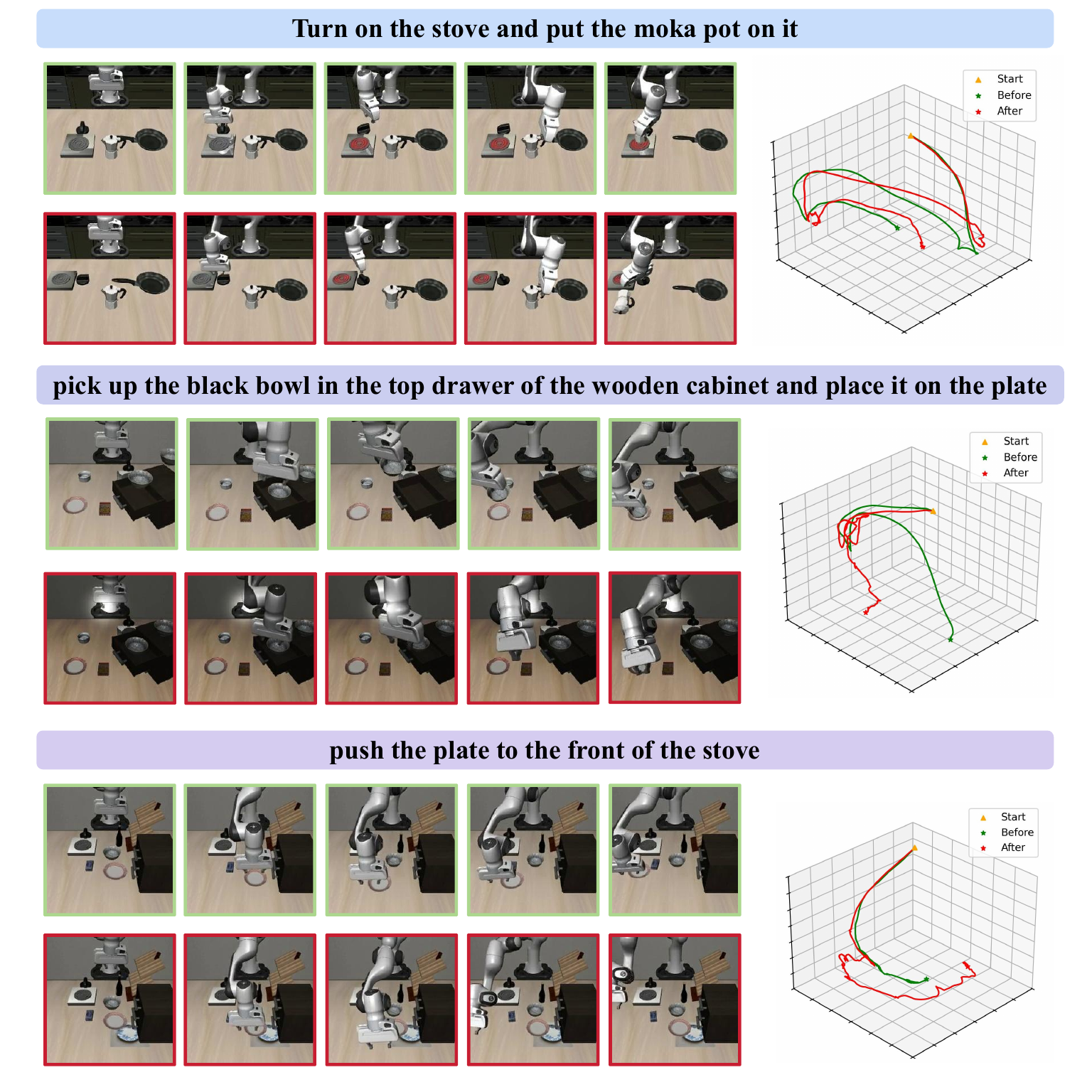}
  \vspace{-2ex}
  \caption{\textbf{Qualitative results under three physical variations on OpenVLA-7B~\cite{kim2024openvla} fine-tuned on LIBERO~\cite{liu2023libero}.} Three manipulation tasks are shown with original executions (top row) and adversarially perturbed executions (bottom row, highlighted in red) for each task. The 3D trajectory visualizations on the right demonstrate the end-effector paths before (\textbf{\textcolor[HTML]{006400}{green}}) and after (\textbf{\textcolor[HTML]{8B0000}{red}}) applying physical variations, illustrating how object 3D transformations (\textbf{top}), illumination variations (\textbf{middle}), and adversarial patches (\textbf{bottom}) respectively disrupt the robot's motion patterns and lead to task failures.}
  \vspace{-3ex}
    \label{fig:experiments}
\end{figure*}

\begin{table*}[t]
\centering
\caption{\textbf{Main Results.} Failure Rate (FR$\uparrow$) of all baselines across different tasks within the LIBERO~\cite{liu2023libero} task suites (Spatial, Object, Goal, and Long). Numbers in \textbf{bold} indicate the best performance. Values in parentheses denote the change relative to the Clean baseline.}
\label{exp:main}
\renewcommand{\arraystretch}{1.1}
\setlength{\tabcolsep}{3.2pt}
\resizebox{\textwidth}{!}{
\begin{tabular}{@{}lcccccccccc@{}}
\toprule
 & \multicolumn{2}{c}{{\small\textbf{Spatial*}}} & \multicolumn{2}{c}{{\small\textbf{Object*}}} & \multicolumn{2}{c}{{\small\textbf{Goal*}}} & \multicolumn{2}{c}{{\small\textbf{Long*}}} & \multicolumn{2}{c}{{\small\textbf{Avg}}} \\
\cmidrule(lr){2-3} \cmidrule(lr){4-5} \cmidrule(lr){6-7} \cmidrule(lr){8-9} \cmidrule(lr){10-11}
{\small\textbf{Type}} & {\small Rand.} & {\small Best} & {\small Rand.} & {\small Best} & {\small Rand.} & {\small Best} & {\small Rand.} & {\small Best} & {\small Rand.} & {\small Best} \\
\midrule
% ======================== OpenVLA ========================
% Clean: S=15.3  O=11.6  G=20.8  L=46.3  A=23.5
\rowcolor{modelAHdr}
\multicolumn{11}{c}{\textbf{\textit{OpenVLA (fine-tuned)}}~\cite{kim2024openvla}} \\
Clean & \multicolumn{2}{c}{15.3} & \multicolumn{2}{c}{11.6} & \multicolumn{2}{c}{20.8} & \multicolumn{2}{c}{46.3} & \multicolumn{2}{c}{23.5} \\
3D Trans. & 48.0\dd{32.7} & \textbf{71.0}\dd{55.7} & 53.0\dd{41.4} & \textbf{79.0}\dd{67.4} & 45.0\dd{24.2} & \textbf{83.0}\dd{62.2} & 77.0\dd{30.7} & \textbf{98.0}\dd{51.7} & 56.0\dd{32.5} & \textbf{83.0}\dd{59.5} \\
Illum. & 28.0\dd{12.7} & \textbf{55.0}\dd{39.7} & 34.0\dd{22.4} & \textbf{48.0}\dd{36.4} & 30.0\dd{9.2} & \textbf{59.0}\dd{38.2} & 60.0\dd{13.7} & \textbf{89.0}\dd{42.7} & 38.0\dd{14.5} & \textbf{63.0}\dd{39.5} \\
Adv. Patch & 23.0\dd{7.7} & \textbf{60.0}\dd{44.7} & 24.0\dd{12.4} & \textbf{62.0}\dd{50.4} & 34.0\dd{13.2} & \textbf{72.0}\dd{51.2} & 52.0\dd{5.7} & \textbf{90.0}\dd{43.7} & 33.0\dd{9.5} & \textbf{71.0}\dd{47.5} \\
\midrule
% ======================== OpenVLA-OFT ========================
% Clean: S=3.8  O=1.7  G=3.8  L=9.3  A=4.7
\rowcolor{modelBHdr}
\multicolumn{11}{c}{\textbf{\textit{OpenVLA-OFT (fine-tuned)}}~\cite{kim2025fine}} \\
Clean & \multicolumn{2}{c}{3.8} & \multicolumn{2}{c}{1.7} & \multicolumn{2}{c}{3.8} & \multicolumn{2}{c}{9.3} & \multicolumn{2}{c}{4.7} \\
3D Trans. & 36.0\dd{32.2} & \textbf{56.0}\dd{52.2} & 38.0\dd{36.3} & \textbf{64.0}\dd{62.3} & 32.0\dd{28.2} & \textbf{67.0}\dd{63.2} & 62.0\dd{52.7} & \textbf{82.0}\dd{72.7} & 42.0\dd{37.3} & \textbf{68.0}\dd{63.3} \\
Illum. & 17.0\dd{13.2} & \textbf{40.0}\dd{36.2} & 21.0\dd{19.3} & \textbf{35.0}\dd{33.3} & 18.0\dd{14.2} & \textbf{44.0}\dd{40.2} & 47.0\dd{37.7} & \textbf{74.0}\dd{64.7} & 26.0\dd{21.3} & \textbf{48.0}\dd{43.3} \\
Adv. Patch & 11.0\dd{7.2} & \textbf{45.0}\dd{41.2} & 13.0\dd{11.3} & \textbf{48.0}\dd{46.3} & 22.0\dd{18.2} & \textbf{58.0}\dd{54.2} & 39.0\dd{29.7} & \textbf{75.0}\dd{65.7} & 21.0\dd{16.3} & \textbf{56.0}\dd{51.3} \\
\midrule
% ======================== UniVLA ========================
% Clean: S=11.0  O=3.0  G=5.0  L=10.0  A=7.3
\rowcolor{modelCHdr}
\multicolumn{11}{c}{\textbf{\textit{UniVLA (fine-tuned)}}~\cite{bu2025univla}} \\
Clean & \multicolumn{2}{c}{11.0} & \multicolumn{2}{c}{3.0} & \multicolumn{2}{c}{5.0} & \multicolumn{2}{c}{10.0} & \multicolumn{2}{c}{7.3} \\
3D Trans. & 17.0\dd{6.0} & \textbf{98.0}\dd{87.0} & 68.0\dd{65.0} & \textbf{98.0}\dd{95.0} & 31.0\dd{26.0} & \textbf{58.0}\dd{53.0} & 82.0\dd{72.0} & \textbf{98.0}\dd{88.0} & 49.5\dd{42.2} & \textbf{88.0}\dd{80.7} \\
Illum. & 5.0\du{6.0} & \textbf{12.0}\dd{1.0} & 17.0\dd{14.0} & \textbf{36.0}\dd{33.0} & 15.0\dd{10.0} & \textbf{40.0}\dd{35.0} & 13.0\dd{3.0} & \textbf{50.0}\dd{40.0} & 12.5\dd{5.2} & \textbf{34.5}\dd{27.2} \\
Adv. Patch & 12.0\dd{1.0} & \textbf{50.0}\dd{39.0} & 30.0\dd{27.0} & \textbf{72.0}\dd{69.0} & 5.0\dd{0.0} & \textbf{60.0}\dd{55.0} & 10.0\dd{0.0} & \textbf{73.0}\dd{63.0} & 14.3\dd{7.0} & \textbf{63.8}\dd{56.5} \\
\midrule
% ======================== pi_0.5 ========================
% Clean: S=2.0  O=3.0  G=3.0  L=8.0  A=4.0
\rowcolor{modelDHdr}
\multicolumn{11}{c}{\textbf{\textit{$\pi_{0.5}$ (fine-tuned)}}~\cite{intelligence2025pi_}} \\
Clean & \multicolumn{2}{c}{2.0} & \multicolumn{2}{c}{3.0} & \multicolumn{2}{c}{3.0} & \multicolumn{2}{c}{8.0} & \multicolumn{2}{c}{4.0} \\
3D Trans. & 18.0\dd{16.0} & \textbf{98.0}\dd{96.0} & 44.0\dd{41.0} & \textbf{88.0}\dd{85.0} & 22.0\dd{19.0} & \textbf{73.0}\dd{70.0} & 54.0\dd{46.0} & \textbf{84.0}\dd{76.0} & 35.0\dd{31.0} & \textbf{86.0}\dd{82.0} \\
Illum. & 4.0\dd{2.0} & \textbf{12.0}\dd{10.0} & 3.0\dd{0.0} & \textbf{11.0}\dd{8.0} & 3.0\dd{0.0} & \textbf{14.0}\dd{11.0} & 8.0\dd{0.0} & \textbf{12.0}\dd{4.0} & 5.0\dd{1.0} & \textbf{12.0}\dd{8.0} \\
Adv. Patch & 17.0\dd{15.0} & \textbf{75.0}\dd{73.0} & 12.0\dd{9.0} & \textbf{43.0}\dd{40.0} & 9.0\dd{6.0} & \textbf{30.0}\dd{27.0} & 10.0\dd{2.0} & \textbf{34.0}\dd{26.0} & 12.0\dd{8.0} & \textbf{46.0}\dd{42.0} \\
\bottomrule

\vspace{-3ex}
\end{tabular}}
\end{table*}

\section{Experiments}

\subsection{Implementation Details}

\noindent\textbf{Object 3D Transformations.} To rigorously stress-test the spatial reasoning capabilities of VLA models and discover the absolute worst-case configurations, we explore the full continuous rotation space. Specifically, we parameterize the 3D transformations across all three axes (yaw, pitch, and roll) to allow objects to assume any arbitrary orientation. This design explicitly simulates chaotic yet physically possible real-world scenarios, such as a knocked-over cup. Optimizing across the complete rotation space provides a more comprehensive assessment of the models' spatial reasoning, helping to ensure that the identified vulnerabilities reflect intrinsic perceptual weaknesses rather than biases introduced by predefined pose limits.

\noindent\textbf{Illumination Variations.} We simulate realistic illumination variations by modeling a dynamic point light source through the optimization of the 4-dimensional parameter space $\boldsymbol{\Lambda}=\{x, y, \sigma, I\}$. The spatial coordinates $(x, y)$ can move freely across the workspace to cast diverse directional shadows, while the spread radius ($\sigma$) and intensity ($I$) are continuously optimized within physically plausible limits. By precluding trivial sensor failures (e.g., complete overexposure or underexposure), this design ensures the evaluation strictly isolates genuine perceptual vulnerabilities induced by complex, deceptive illumination patterns.

\noindent\textbf{Adversarial Patch Placement.}
Rather than optimizing uninterpretable, pixel-level patch textures~\cite{wang2024exploring}, we employ benign natural images and strategically optimize their spatial placement across the tabletop. By integrating these patches into the environment via standard UV mapping, our approach ensures strict physical realizability. Since the table occupies a dominant portion of the robot's visual field, these surface-level perturbations can effectively disrupt the spatial context and misguide the model's attention without physically occluding the target objects. Formally, we define a 2-dimensional spatial vector $\boldsymbol{\phi} = \{x, y\} \in [\boldsymbol{\phi}_{\min}, \boldsymbol{\phi}_{\max}]$, where $(x, y)$ denotes the placement coordinates of the adversarial patch within the table's texture map. Here, the boundaries $\boldsymbol{\phi}_{\min}$ and $\boldsymbol{\phi}_{\max}$ are defined exclusively by the natural physical dimensions of the workspace surface. This formulation avoids arbitrary search constraints, ensuring the patch remains a valid background feature while allowing the optimization algorithm to discover the most severe adversarial configurations anywhere on the table.

\noindent\textbf{Baselines.}
To comprehensively evaluate the effectiveness of our methods, we establish three baseline conditions for comparison. First, we assess models on the clean environment, which serves as the control condition where models operate under normal circumstances without any adversarial modifications. Second, we implement a random strategy that samples transformation parameters uniformly from the initial parameter spaces without optimization, quantifying the impact of unguided perturbations and demonstrating the necessity of our optimization-based approach over naive random exploration. Third, we evaluate the best-case attacks using optimized parameters obtained through our proposed optimization framework, representing the most effective parameters discovered for each physical transformation type.

\subsection{Experiment Setup}
\noindent\textbf{Dataset \& Threat Models.}
We conduct our experiments exclusively on the LIBERO dataset~\cite{liu2023libero} in simulation environments, which is a comprehensive simulation dataset designed to evaluate vision-language-action models across four distinct task categories: Spatial, Object, Goal, and Long-horizon tasks. Our method involves physical transformations that require precise control and reproducibility of environmental conditions. Simulation environments enable us to systematically apply and evaluate these physical transformations under controlled conditions, ensuring that the attacks are repeatable and measurable. We select publicly available and state-of-the-art VLAs as victim models for a comprehensive evaluation. Specifically, our evaluation encompasses a diverse suite of cutting-edge architectures: four baseline OpenVLA~\cite{kim2024openvla} variants fine-tuned on distinct LIBERO task suites (Spatial, Object, Goal, and Long), their four corresponding OpenVLA-OFT~\cite{kim2025fine} variants, four UniVLA~\cite{bu2025univla} variants identically fine-tuned across the respective LIBERO suites, and the LIBERO-finetuned $\pi_{0.5}$~\cite{intelligence2025pi_} model. This extensive selection ensures a rigorous assessment of our adversarial framework's effectiveness across different model capacities, architectures, and fine-tuning strategies.

\noindent\textbf{Evaluation Metric.}
For task execution assessment, we employ the maximum step count from each LIBERO task suite's training data as the termination criterion for timeout failures. We utilize the Failure Rate (FR), calculated as 1$-$Success Rate (SR), as established in LIBERO~\cite{liu2023libero}, as our principal performance indicator.
% For task execution assessment, we employ the maximum step count from each LIBERO task suite's training data as the termination criterion for timeout failures, thereby minimizing computational costs. We utilize the Failure Rate (FR), calculated as 1$-$Success Rate (SR), as established in LIBERO~\cite{liu2023libero}, as our principal performance indicator. In the context of robustness evaluation, emphasizing FR over SR provides a more direct and intuitive quantification of model vulnerability, explicitly highlighting the effectiveness of our proposed method in disrupting VLA policies.

\noindent\textbf{Physical Experiment setting.}
We adopt a robotic platform consisting of an AgileX Piper arm equipped with a 1-DOF gripper, providing 7 degrees of freedom (DoF) for manipulation tasks. The visual sensing system comprises two RealSense D435if cameras: one mounted in a fixed external position to capture third-person observations, and another mounted on the robot's wrist to collect localized wrist-view images.

\noindent\textbf{Evaluation Details.}
To evaluate the effectiveness of our proposed methods, we conduct experiments on the LIBERO dataset ~\cite{liu2023libero} following the evaluation protocol established by Kim et al.~\cite{kim2024openvla}. Each task suite comprises 10 distinct tasks, with 50 trials performed per task, yielding 500 total rollouts per suite. To balance computational efficiency with attack effectiveness, we configure our optimization process with 10 samples per iteration across 50 iterations, enabling thorough exploration of the adversarial parameter space while maintaining reasonable computational overhead. All experiments are performed on a single NVIDIA A800 GPU with 80GB memory.

\begin{figure*}[t]
  \centering
  \includegraphics[width=\linewidth]{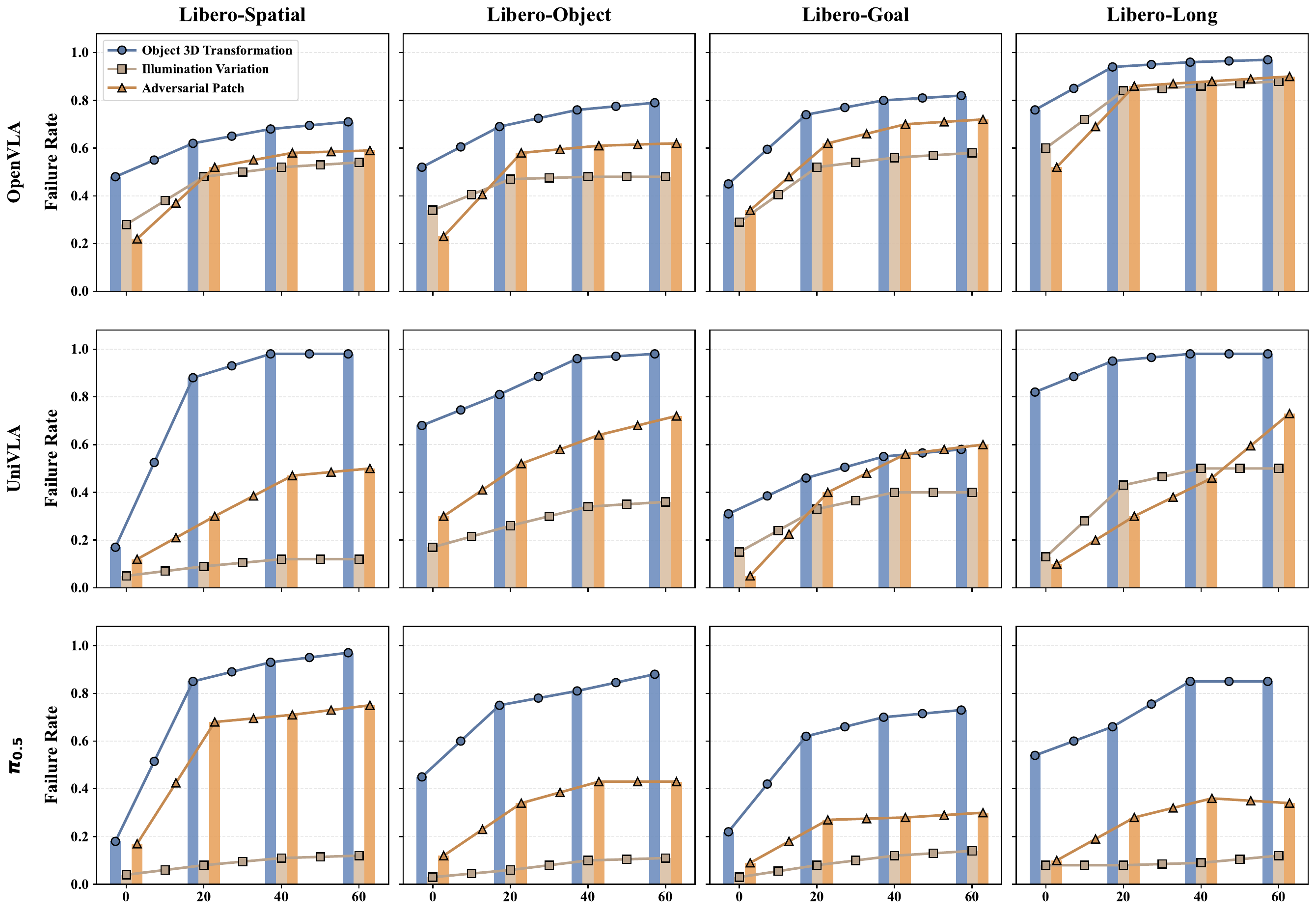}
  % \vspace{-2ex}
    \caption{\textbf{Ablation study on optimization steps.} We evaluate task failure rates under three adversarial conditions across four LIBERO task suites (Spatial, Object, Goal, and Long). The x-axis represents optimization iterations (0-60), and the y-axis shows the failure rate. The overlaid lines demonstrate the increasing trend of failure rates as the optimization progresses.} 

  \vspace{-3ex}
    \label{fig:ablation study: step}
\end{figure*}

\subsection{Main Results}
\noindent\textbf{Quantitative Results.} Table~\ref{exp:main} presents the failure rates of the evaluated models (OpenVLA, OpenVLA-OFT, UniVLA, and $\pi_{0.5}$) under various physical transformations across four LIBERO task suites. Beyond simply reporting performance metrics, these results validate our framework's capacity to systematically uncover alarming vulnerabilities in VLA architectures. While these models demonstrate strong capabilities in controlled settings (with clean failure rates ranging from 4.0\% to 23.5\%), this laboratory success fails to translate to physical robustness. By formulating discrete physical variations as continuous optimization problems, our mechanism successfully discovers severe worst-case scenarios. Specifically, 3D object transformations ($\boldsymbol{\Theta}$), which heavily challenge spatial reasoning, consistently emerge as the most devastating threat, driving the average failure rate of OpenVLA to 83.0\% and causing even the highly capable UniVLA to surge to an 88.0\% failure rate. Similarly, variations targeting visual perception via illumination changes ($\boldsymbol{\Lambda}$) and scene understanding via adversarial patches ($\boldsymbol{\phi}$) inflict substantial degradation across almost all evaluated models. \textit{\textbf{Notably, even the state-of-the-art $\pi_{0.5}$ model (4.0\% clean rate) completely collapses under $\boldsymbol{\Theta}$ attacks with an 86.0\% failure rate.}} Crucially, the data highlights the necessity of our optimization approach: while random environmental perturbations cause notable performance drops (e.g., 33.0\%-56.0\% for OpenVLA, 21.0\%-42.0\% for OpenVLA-OFT), our targeted framework efficiently discovers the extreme bounds of VLA brittleness. This vulnerability is especially pronounced in long-horizon tasks, where failure rates approach near-total collapse under optimized object transformations (e.g., for OpenVLA and UniVLA), indicating that adversarial effects compound severely over extended action sequences, while simpler Object and Spatial tasks show relatively lower but still substantial vulnerability. These consistent patterns expose a critical gap between current VLA capabilities and their readiness for unpredictable real-world deployments.

\noindent\textbf{Qualitative Results.} We qualitatively analyze robot movement trajectories under the three proposed physical variation types in Fig.~\ref{fig:experiments}. For object 3D transformations, as shown in the trajectory plots, the robot maintains similar initial movement patterns but fails to achieve correct object placement due to spatial misalignment. The trajectory demonstrates that while the robot attempts to complete the full task sequence, the geometric perturbations cause mislocalization of target positions, resulting in the robot placing objects at incorrect locations. We attribute this to the disruption of the model's spatial reasoning capabilities through geometric transformations, causing systematic errors in perceiving and reaching intended placement positions. For illumination variations, we observe degraded object recognition that manifests as incomplete grasping actions and premature trajectory termination. The adversarial lighting conditions interfere with the model's visual perception, leading to failed object detection and misaligned manipulation attempts. For adversarial patches, we observe oscillatory behaviors and intermittent loss of object contact. The patches induce consistent directional biases in the robot's movements, causing it to deviate from intended pushing trajectories and lose task-relevant object interactions. In summary, our qualitative analysis demonstrates that all three physical variations can effectively disrupt robot actions through distinct failure modes. Additional visualization results are provided in Appendix B.

\begin{figure*}[t]
  \centering
  \includegraphics[width=\linewidth]{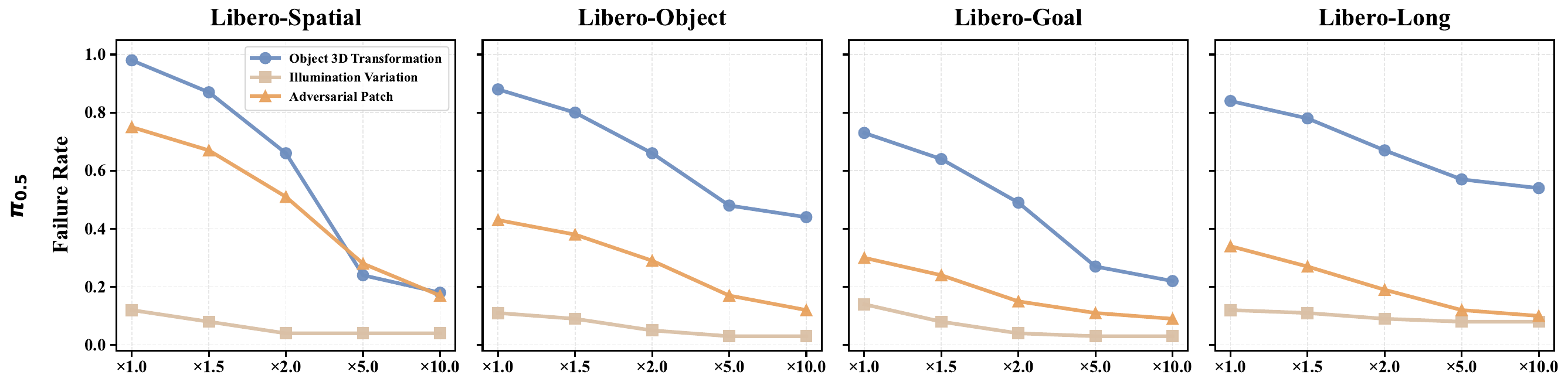}
  % \vspace{-2ex}
  \caption{\textbf{Ablation study on the scale ratio of optimal distribution.} We evaluate task failure rates under three adversarial conditions across four LIBERO task suites (Spatial, Object, Goal, and Long). The x-axis represents the scale ratio of optimal $\Sigma$, and the y-axis shows the failure rate.}
  \vspace{-3ex}
    \label{fig:ablation study:scale ratio}
\end{figure*}

\begin{figure*}[t]
  \centering
  \includegraphics[width=\linewidth]{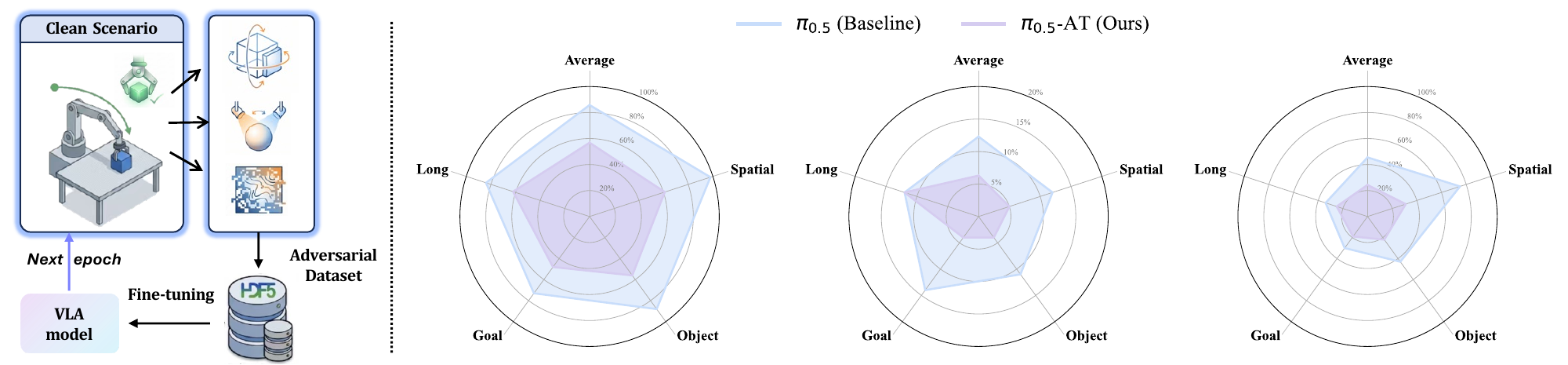}
  % \vspace{-2ex}
    \caption{\textbf{Adversarial Training Results.} \textbf{(Left)} Schematic diagram of the proposed adversarial training pipeline. \textbf{(Right)} Worst-case Failure Rate (FR) comparisons (from left to right: Object 3D Transformations, Illumination Variations, and Adversarial Patches) between the baseline $\pi_{0.5}$ and the improved $\pi_{0.5}$-AT across the LIBERO suite.}
  \vspace{-3ex}
    \label{fig:ablation study:radar}
\end{figure*}

\subsection{Ablation Study and Additional Results}

\noindent\textbf{Optimization iterations.} We conduct comprehensive ablation experiments analyzing the effectiveness of our continuous optimization mechanism across varying iteration steps (0 to 60) on four LIBERO task suites, as shown in Fig.~\ref{fig:ablation study: step}. Across the evaluated models (OpenVLA, UniVLA, and $\pi_{0.5}$), all physical variations exhibit a distinct, rapid convergence pattern. The optimization mechanism efficiently discovers severe vulnerabilities within the first 40 iterations, where failure rates surge sharply from their random initial states before stabilizing. For instance, under Object 3D Transformations on the Spatial task, the failure rate for $\pi_{0.5}$ spikes dramatically from 18.0\% at initialization to 93.0\% at 40 steps. Similarly, UniVLA's failure rate reaches an overwhelming 98.0\% by iteration 40. This trend underscores the efficiency of our black-box optimization in systematically locating worst-case physical states without requiring prohibitive computational overhead. \textit{\textbf{These findings validate our premise: simply applying random physical perturbations (iteration 0) is insufficient to evaluate model robustness.}} Continuous optimization is strictly necessary to expose the true boundaries of VLA brittleness, particularly as adversarial effects amplify during complex manipulation tasks.

\noindent\textbf{Scaling ratios of the optimal distribution.} To further validate that the vulnerabilities uncovered by our framework are highly specific worst-case scenarios rather than general sensitivities to arbitrary noise, we analyze the impact of scaling the optimal perturbation distribution in Fig.~\ref{fig:ablation study:scale ratio}. Using the highly capable $\pi_{0.5}$ model as a case study, we apply scaling multipliers ($\times1.0$ to $\times10.0$) to the optimal distributions discovered by our framework. The data reveals a clear inverse relationship. \textit{\textbf{As the scaling ratio increases, the optimal distribution becomes more diffuse and unstructured, causing a substantial drop in attack effectiveness.}} For example, under Object 3D Transformations on the Spatial task, the failure rate steadily decreases from 98.0\% at the precise optimal distribution ($\times1.0$) down to 18.0\% when scaled by $\times10.0$. A similar degradation in attack potency occurs with adversarial patches, where the failure rate drops from 75.0\% down to 17.0\% under the same scaling conditions. This significant decline confirms that VLA models do not simply fail due to general environmental variations; rather, they are susceptible to specific, critical geometric and visual configurations.

% This firmly highlights the necessity of the Eva-VLA framework: precise formulation and continuous optimization are essential to effectively locate the exact worst-case scenarios that threaten real-world robotic deployments.

\noindent\textbf{Defense Mechanisms and Robustness Improvements.} We demonstrate that the worst-case scenarios generated by Eva-VLA can also provide meaningful guidance for robustness improvements. By collecting adversarial examples of the specific physical variations, we conduct adversarial training to enhance the model's adversarial robustness against these targeted threats. As illustrated in the radar charts (Fig.~\ref{fig:ablation study:radar}), applying this adversarial training to the $\pi_{0.5}$ model explicitly reduces its vulnerability to optimized adversarial patches ($\boldsymbol{\phi}$) from an average failure rate of 45.5\% to 24.3\%, and mitigates illumination variations ($\boldsymbol{\Lambda}$) from 12.3\% to 6.3\%. Furthermore, against the highly challenging object 3D transformations ($\boldsymbol{\Theta}$), the failure rate drops notably from 85.8\% to 56.8\%. \textit{\textbf{Crucially, this substantial enhancement in robustness comes at a negligible cost to standard performance, with the average clean task failure rate experiencing only a marginal increase from 4.0\% to 5.0\%.}} These results confirm that the physical perturbations discovered by Eva-VLA capture critical, learnable architectural flaws rather than arbitrary uninterpretable noise. Detailed information regarding the adversarial data composition and fine-tuning hyperparameters is provided in Appendix C.

\begin{figure}[t]
  \centering
  \vspace{-2ex}
  \includegraphics[width=\linewidth]{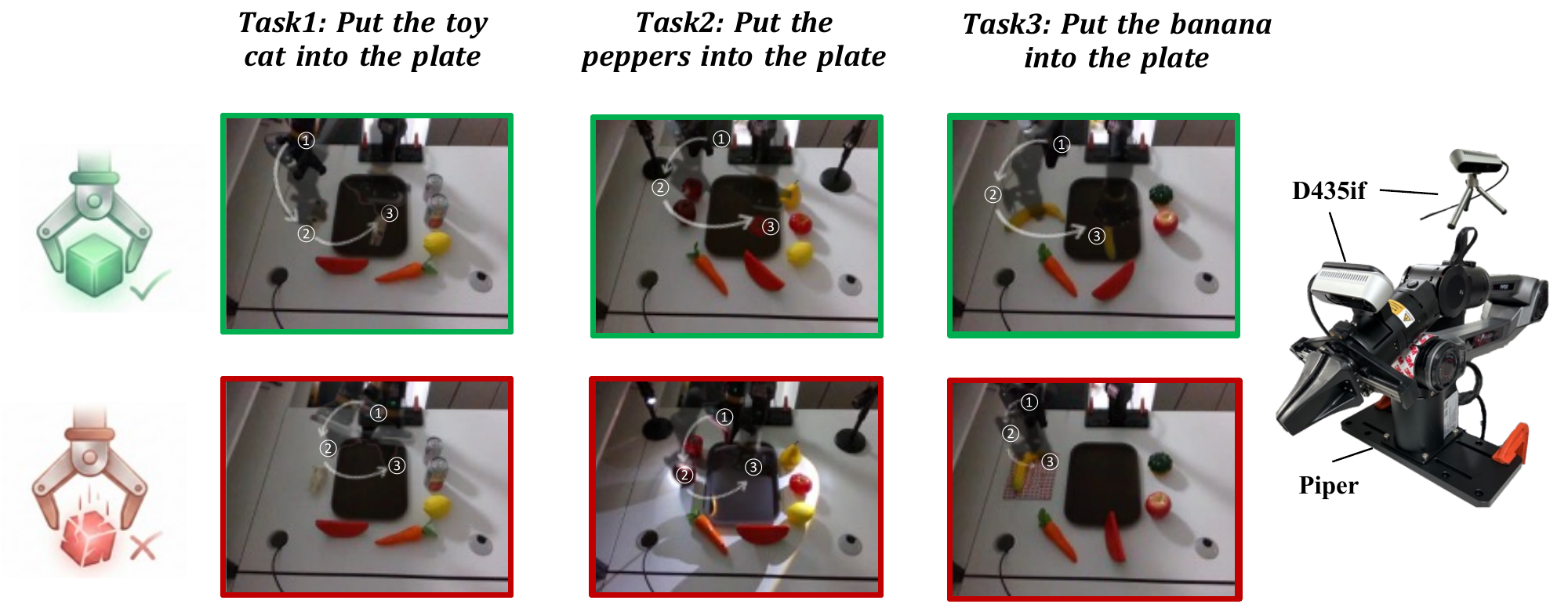}
  \vspace{-3ex}
    \caption{\textbf{Physical Experiments.} \textbf{(Left)} Qualitative results in the real world. The top row demonstrates successful task executions in clean environments. The bottom row illustrates corresponding failure cases induced by the three proposed physical variations: object 3D transformations, illumination variations, and adversarial patches. \textbf{(Right)} The experimental hardware setup, which consists of an AgileX Piper robotic arm equipped with a 1-DOF gripper and a RealSense D435if camera for capturing third-person observations.}
  \vspace{-2ex}
    \label{fig:physical}
\end{figure}

\noindent\textbf{Real-World Performance.} Beyond the simulation results, we conduct comprehensive evaluations of the three physical variations in real-world scenarios. Specifically, we focus on the representative ``\textit{put A into B}'' instruction paradigm, designing three distinct tasks tailored to evaluate each variation type independently. To establish a robust baseline, we first collect 50 clean trajectories for each task. As demonstrated in Fig.~\ref{fig:physical}, all three variations induce severe task failures: applying 3D transformations to critical objects disrupts spatial manipulation, extreme illumination variations obscure visual perception, and adversarial patch placements severely interfere with scene understanding. Moreover, under these perturbations, the robot frequently exhibits unstable, jerky, and oscillatory motions consistent with our simulation observations, posing significant risks to human safety and the operational environment. Additional detailed results are provided in the Appendix D.

\section{Conclusion}
In this paper, we propose Eva-VLA, a novel framework that systematically investigates the vulnerabilities of VLA models, uncovering significant performance degradation under physical variations. Extensive experiments demonstrate that even state-of-the-art VLA models are highly susceptible to these challenges, with issues persisting in simulated and real-world settings. The observed instability and failures in physical experiments highlight potential risks to operational safety, underscoring the urgent need for enhanced defenses against physical perturbations. These findings provide critical insights for improving the robustness of VLA-based robotic systems, offering a viable pathway toward their safe and reliable deployment in physical environments.

\clearpage  % TODO FINAL: This \clearpage needs to be removed from both review and camera-ready versions.

\bibliographystyle{splncs04}
\bibliography{main}
\end{document}

\title{The Path to Reconciling Quality and Safety in Text-to-Image Generation: Dataset, Method, and Evaluation}

% TODO REVIEW: If the paper title is too long for the running head
\titlerunning{Reconciling Quality and Safety in T2I Generation}

\author{First Author\inst{1} \and
Second Author\inst{2}}

\authorrunning{F.~Author et al.}

\institute{Institution1, Address1 \\
\email{firstauthor@i1.org} \and
Institution2, Address2 \\
\email{secondauthor@i2.org}}

% \maketitle

\clearpage
\setcounter{section}{0}
\renewcommand{\thesection}{\Alph{section}}

\definecolor{colorHate}{HTML}{C0392B}      % Deep Red
\definecolor{colorHarass}{HTML}{8E44AD}    % Purple
\definecolor{colorViolence}{HTML}{D35400}  % Pumpkin/Burnt Orange
\definecolor{colorSelfHarm}{HTML}{16A085}  % Teal
\definecolor{colorSexual}{HTML}{D81B60}    % Pink/Magenta
\definecolor{colorShock}{HTML}{7F8C8D}     % Grey/Slate (or Olive)
\definecolor{colorIllegal}{HTML}{2C3E50}   % Midnight Blue

% --- Define the Custom TColorBox Environment ---
\newtcolorbox{rubricbox}[2]{
    enhanced,
    colframe=#1,           % Border color
    colback=#1!5!white,    % Background color (very light version of border)
    coltitle=white,        % Title text color
    title={\textbf{#2}},   % Title content
    fonttitle=\large\bfseries,
    arc=2mm,               % Rounded corners radius
    boxrule=1.2pt,         % Border thickness
    left=2mm, right=2mm, top=2mm, bottom=2mm,
    toptitle=1mm, bottomtitle=1mm,
    drop shadow,     % Subtle shadow for depth
    breakable              % Allows box to split across pages if necessary
}

% --- Custom Itemize for the Rubric Levels ---
\newlist{rubriclist}{itemize}{1}
\setlist[rubriclist]{
    label={}, 
    leftmargin=0pt, 
    itemindent=0pt, 
    listparindent=0pt,
    parsep=4pt,
    itemsep=4pt
}
\newcommand{\level}[2]{\item \textbf{\textcolor{black}{#1}}: #2}

% =================================================================
% APPENDIX CONTENT START
% =================================================================

\section*{Appendix A. Algorithmic Details of the Optimization Framework}
\label{appendix:algorithm}

In this section, we provide the complete algorithmic details of the continuous black-box optimization framework utilized in \textbf{\textit{Eva-VLA}}. As outlined in the main text, discovering the worst-case physical variations is formulated as a gradient-free optimization problem and solved using the Covariance Matrix Adaptation Evolution Strategy (CMA-ES).

Algorithm~\ref{alg:optimization} details the complete iterative procedure. The process aims to find the optimal physical transformation parameters $\boldsymbol{\mathcal{T}}^*$ that maximize the adversarial loss $\mathcal{L}_{adv}$ against the Vision-Language-Action (VLA) model $F$. 

\begin{algorithm}[h] 
\small
\caption{\small Eva-VLA Optimization Algorithm}\label{alg:optimization}
\KwData{Task dataset $\mathcal{D} = \{(X_j, Y_j, A_j)\}_{j=1}^N$, VLA model $F$, Attack objective $\mathcal{L}_{adv}$, Transformation function $T$, Maximum iterations $t_{max}$, Population size $K$.}
\KwResult{Optimal adversarial transformation configuration $\boldsymbol{\mathcal{T}}^*$.}
\tcc{\scriptsize Initialization of distribution parameters}
$t \leftarrow 0$, $\boldsymbol{\mu}_0 \leftarrow \mathbf{0}$, $\boldsymbol{C}_0 \leftarrow \mathbf{I}$, $\boldsymbol{\Sigma}_0 \leftarrow 0.1$, $\mathbf{p} \leftarrow \mathbf{0}$\;

\While{$t < t_{\max}$}{
    \tcc{\scriptsize Step 1: Sample and evaluate candidates}
    \For{$i = 1 \to K$}{
        $\boldsymbol{\mathcal{T}}_i^{t+1} \sim \mathcal{N}(\boldsymbol{\mu}_t, \boldsymbol{C}_t \boldsymbol{\Sigma}_t^2)$\;
        
        \tcc{\scriptsize Execute task in simulation and get adversarial action}
        $(A_{adv})_i^{t+1} \leftarrow F(T(\boldsymbol{\mathcal{T}}_i^{t+1},X),Y)$\;
        
        \tcc{\scriptsize Calculate loss value (including failure penalty)} 
        $\mathcal{L}_i^{t+1} = \mathcal{L}_{adv}((A_{adv})_i^{t+1}, A_{clean}) + \lambda \cdot \mathbb{I}_{fail}$\;
    }
    
    \tcc{\scriptsize Step 2: Selection based on adversarial effectiveness}
    Sort candidates such that $\mathcal{L}_1^{t+1} \geq \mathcal{L}_2^{t+1} \geq \dots \geq \mathcal{L}_K^{t+1}$\;
    
    \tcc{\scriptsize Step 3: Update Distribution Parameters}
    $\boldsymbol{\mu}_{t+1} \leftarrow \sum_{i=1}^{K} w_i \boldsymbol{\mathcal{T}}_i^{t+1}$\;
    
    \tcc{\scriptsize Update evolution path and covariance matrix}
    $\mathbf{p} \leftarrow (1-c_\sigma)\mathbf{p} + \sqrt{c_\sigma(2-c_\sigma)\mu_{eff}} \boldsymbol{C}_t^{-\frac{1}{2}} \frac{\boldsymbol{\mu}_{t+1} - \boldsymbol{\mu}_t}{\boldsymbol{\Sigma}_t}$\;
    $\boldsymbol{C}_{t+1} \leftarrow (1 - c_c) \boldsymbol{C}_t + c_c \sum_{i=1}^{K} w_i \left(\frac{\boldsymbol{\mathcal{T}}_i^{t+1} - \boldsymbol{\mu}_t}{\boldsymbol{\Sigma}_t}\right)\left(\frac{\boldsymbol{\mathcal{T}}_i^{t+1} - \boldsymbol{\mu}_t}{\boldsymbol{\Sigma}_t}\right)^T$\;
    
    \tcc{\scriptsize Step 4: Learning Rate Adaptation (LRA)}
    $\boldsymbol{\Sigma}_{t+1} \leftarrow \boldsymbol{\Sigma}_t \cdot \exp\left( \frac{c_\sigma}{d_\sigma} \left( \frac{\|\mathbf{p}\|}{\mathbb{E}[\|\mathcal{N}(0, I)\|]} - 1 \right) \right)$\;
    
    \tcc{\scriptsize Step 5: Early Stopping Policy}
    \If{$|\boldsymbol{\mu}_{t+1} - \boldsymbol{\mu}_t| < \epsilon$ \textbf{or} improvements plateau}{
        \textbf{break}\;
    }
    $t \leftarrow t + 1$\;
}
$\boldsymbol{\mu}^* \leftarrow \boldsymbol{\mu}_{t}, \boldsymbol{C}^* \leftarrow \boldsymbol{C}_{t}, \boldsymbol{\Sigma}^* \leftarrow \boldsymbol{\Sigma}_{t}$\;
$\boldsymbol{\mathcal{T}}^* \sim \mathcal{N}(\boldsymbol{\mu}^*, \boldsymbol{C}^* (\boldsymbol{\Sigma}^*)^2)$\;
\end{algorithm}

\vspace{1ex}
\noindent\textbf{Detailed Step Description:}
\begin{enumerate}
    \item \textbf{Initialization:} The algorithm initializes the mean vector $\boldsymbol{\mu}_0$, the covariance matrix $\boldsymbol{C}_0$ as an identity matrix, the step size $\boldsymbol{\Sigma}_0$ (acting as the baseline learning rate), and the evolution path $\mathbf{p}$.
    \item \textbf{Sampling \& Evaluation (Step 1):} In each generation, a population of $K$ candidate transformations is sampled. The transformations are applied to the simulation environment via $T(\cdot)$, and the VLA model $F$ is queried. The adversarial loss $\mathcal{L}$ is computed based on the deviation from the clean action $A_{clean}$, augmented by a penalty term $\lambda \cdot \mathbb{I}_{fail}$ indicating whether a complete task failure occurred.
    \item \textbf{Selection \& Update (Steps 2 \& 3):} Candidates are sorted in descending order of their loss (to maximize disruption). The distribution mean $\boldsymbol{\mu}$ and covariance matrix $\boldsymbol{C}$ are updated using a weighted sum of the most successful candidates ($w_i$). The evolution path $\mathbf{p}$ captures the trajectory of consecutive updates.
    \item \textbf{Learning Rate Adaptation \& Early Stopping (Steps 4 \& 5):} The global step size $\boldsymbol{\Sigma}$ is dynamically adjusted by comparing the length of the evolution path $\|\mathbf{p}\|$ to its expected length under a random walk ($\mathbb{E}[\|\mathcal{N}(0, I)\|]$). If the algorithm detects convergence (i.e., variations between generations fall below a threshold $\epsilon$), the early stopping policy terminates the loop to preserve computational resources.
\end{enumerate}

\noindent\textbf{Default Hyperparameters.} To ensure evaluation reproducibility and avoid manual tuning bias, all internal hyperparameters of the CMA-ES algorithm are set to their standard heuristic defaults derived from the search space dimensionality $D$~\cite{hansen2016cma}. Specifically, the recombination weights ($w_i$), the effective variance selection mass ($\mu_{eff}$), the learning rates for the covariance matrix update ($c_c$) and the step-size update ($c_\sigma$), as well as the damping factor ($d_\sigma$), are automatically configured based on standard CMA-ES implementations.

\noindent\textbf{Computational Overhead and Runtime Analysis.} 
To demonstrate the practical efficiency of our optimization framework, we analyze the computational overhead incurred during the adversarial generation process. All optimization experiments were conducted on a workstation equipped with a single 48GB NVIDIA GPU. 

The overall runtime of Eva-VLA is primarily bottlenecked by the forward inference speed of the specific VLA model being evaluated, whereas the time consumed by physical simulator rendering and CMA-ES parameter updating is comparatively negligible. As detailed in Table~\ref{tab:runtime}, the computational cost varies across different model architectures. For instance, the pi0.5 model requires approximately 10 seconds for a single forward pass, while OpenVLA, OpenVLA-OFT, and UniVLA exhibit longer inference times ranging from 15 to 30 seconds per sample. Assuming a standard CMA-ES population size of $K=10$, each optimization generation takes roughly 1.5 to 5 minutes depending on the target model. Aided by the Early Stopping policy, the framework effectively bounds the total number of generations, ensuring that discovering worst-case variations remains highly tractable even for computationally heavy VLA models.

\begin{table}[h]
\centering
\renewcommand{\arraystretch}{1.2} % 稍微加一点行高，配合换行的表头更好看
\caption{Average computational time for different VLA models during the Eva-VLA optimization process.}
\label{tab:runtime}
\begin{tabular}{lcc}
\toprule
\textbf{Model} & \textbf{Time per Sample (s)} & \textbf{Est.Time per Gen (min)} \\
\midrule
pi0.5~\cite{intelligence2025pi_}        & $\sim$ 10 & $\sim$ 1.6 \\
OpenVLA~\cite{kim2024openvla}      & $\sim$ 15  & 4.1 -- 5.0 \\
OpenVLA-OFT~\cite{kim2025fine}  & 25 -- 30  & 4.1 -- 5.0 \\
UniVLA~\cite{bu2025univla}       & 25 -- 30  & 4.1 -- 5.0 \\
\bottomrule
\end{tabular}
\end{table}

\noindent\textbf{More Ablation Study on Optimization Components.} To validate the efficacy of the core components in our continuous optimization framework, we conduct ablation studies focusing on the Learning Rate Adaptation (LRA) mechanism and the Early Stopping policy. We evaluate these components specifically on the object 3D transformations (pose variations) within the LIBERO-Spatial suite, setting the maximum number of optimization generations to $50$.

\begin{table}[h]
\centering
\caption{Ablation study on Learning Rate Adaptation (LRA) for object 3D transformations on the LIBERO-Spatial suite. We report the Failure Rate (FR) and the average generations required to converge (with a maximum limit of $50$ generations).}
\label{tab:ablation_lra}
\begin{tabular}{lcc}
\toprule
\textbf{Configuration} & \textbf{Failure Rate (FR) $\uparrow$} & \textbf{Avg. Generations $\downarrow$} \\
\midrule
Fixed $\boldsymbol{\Sigma}$ (Small)  & 34.0\% & 50 \\
Fixed $\boldsymbol{\Sigma}$ (Medium) & 82.0\% & 42 \\
Fixed $\boldsymbol{\Sigma}$ (Large)  & 38.0\% & \textbf{18} \\

\midrule
\textbf{Ours (w/ LRA)}               & \textbf{98.0\%} & 34 \\
\bottomrule
\end{tabular}
\end{table}

\noindent\textbf{Learning Rate Adaptation (LRA).} LRA is designed to dynamically adjust the covariance step size $\boldsymbol{\Sigma}$ during optimization, effectively balancing global exploration and local exploitation. To demonstrate its necessity, we compare our full Eva-VLA framework against variants utilizing fixed step sizes (Small, Medium, and Large). As shown in Table~\ref{tab:ablation_lra}, employing a fixed small step size restricts the search capability, causing the optimization to frequently exhaust the maximum $50$ generations while getting trapped in local optima, yielding a low Failure Rate (FR) of 34.0\%. Conversely, a fixed large step size causes severe search oscillation and divergence, terminating early (average 18 generations) but completely failing to pinpoint the worst-case variations (38.0\% FR). While the medium step size performs reasonably well, incorporating LRA enables our framework to dynamically adapt to the non-differentiable physical simulation space. Consequently, our full method achieves the highest FR of 98.0\% in a highly efficient manner (averaging only 34 generations).

\begin{table}[h]
\centering
\renewcommand{\arraystretch}{1.2}
\caption{Ablation study on the Early Stopping Policy for object 3D transformations (pose variations) on the LIBERO-Spatial suite. We compare the computational cost (average generations and time per task) and the attack performance (Failure Rate) against fixed-generation baselines.}
\label{tab:ablation_early_stopping}
% 将整个表格缩放到 \textwidth，保持长宽比 (!)
\resizebox{\textwidth}{!}{
\begin{tabular}{lccc}
\toprule
\textbf{Configuration} & \textbf{Failure Rate (FR) $\uparrow$} & \textbf{Avg. Generations $\downarrow$} & \textbf{Time per Task (min) $\downarrow$} \\
\midrule
Fixed Gen ($t_{max} = 50$)   & \textbf{100.0\%} & 50 & $\sim$ 180.0 \\
Fixed Gen ($t_{max} = 100$)  & \textbf{100.0\%} & 100 & $\sim$ 360.0 \\
\midrule
\textbf{Ours (w/ Early Stop)} & 98.0\% & \textbf{34} & \textbf{$\sim$ 100.0} \\
\bottomrule
\end{tabular}
}
\end{table}

\noindent\textbf{Early Stopping Policy.} The Early Stopping mechanism aims to terminate the continuous optimization process when the adversarial objective plateaus, thereby mitigating unnecessary computational overhead in the physical simulator. To evaluate its impact, we compare our dynamic framework against baselines forced to execute for a fixed number of maximum generations ($t_{max} = 50$ and $100$) specifically on the object 3D transformations (pose variations) within the LIBERO-Spatial suite. 

As detailed in Table~\ref{tab:ablation_early_stopping}, running the optimization for a fixed, large number of generations (e.g., $50$ or $100$) guarantees a perfect 100.0\% Failure Rate (FR). However, this brute-force approach yields no marginal improvements during the later stages of optimization, drastically inflating the computational cost to as much as $\sim$360 minutes per task. By contrast, our Early Stopping policy autonomously truncates redundant searches based on loss stagnation. This intelligent truncation reduces the average required generations to 34, cutting the execution time down to $\sim$100 minutes per task. Crucially, this massive gain in computational efficiency comes at a negligible cost to the attack's effectiveness, strictly preserving a highly severe 98.0\% FR.

\section*{Appendix B. Additional Qualitative Results on the LIBERO Benchmark}
\label{appendix:visualizations}

In this section, we provide extended qualitative results and visualizations for the $\pi_{0.5}$~\cite{intelligence2025pi_} model to further illustrate the severe impact of our proposed physical variations on VLA models. While the main text highlights representative failure modes, the visual examples presented here encompass a broader range of manipulation tasks across the LIBERO~\cite{liu2023libero} benchmark suites (including Spatial, Object, Goal, and Long). 

\begin{figure*}[p]
    \centering
    \includegraphics[page=1, width=\textwidth]{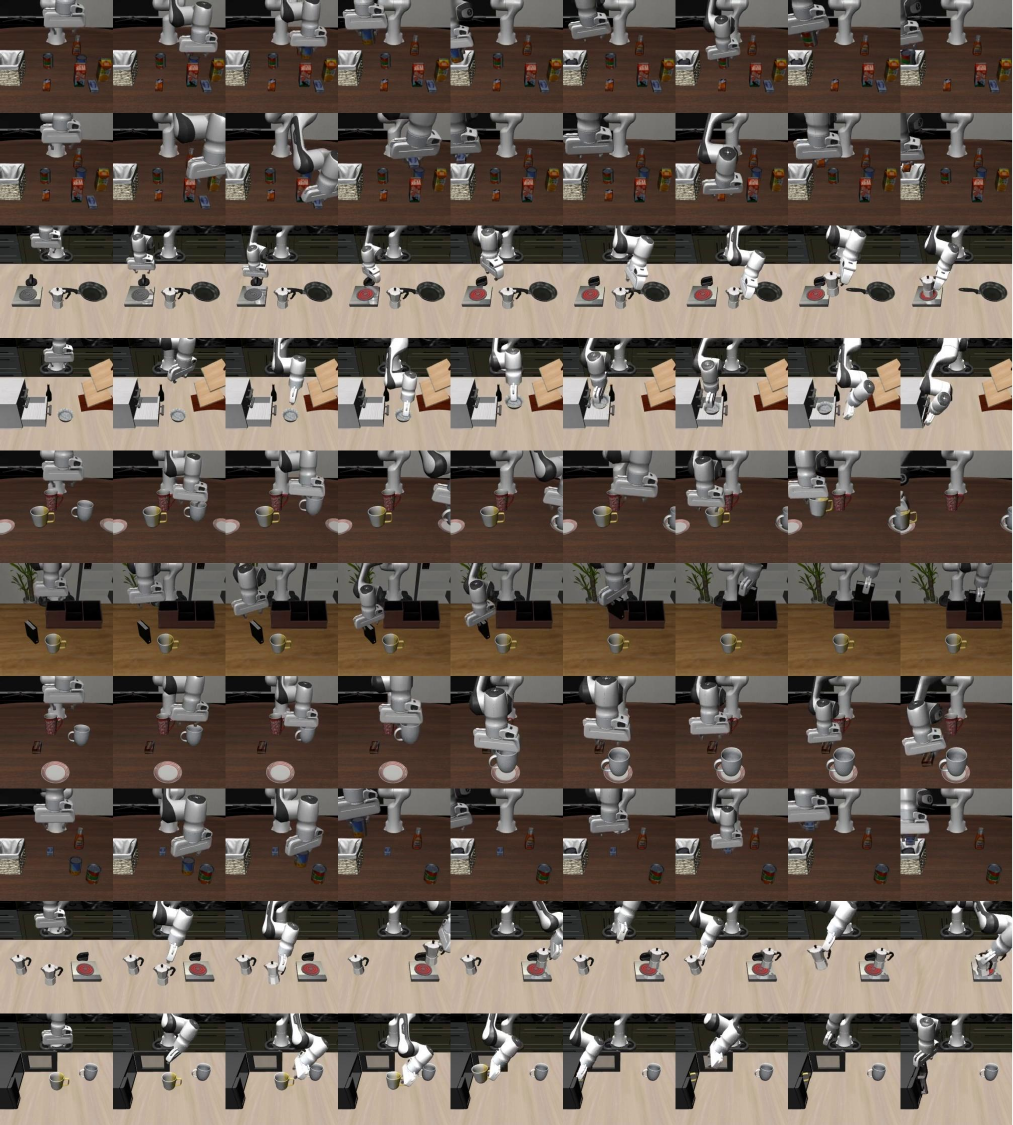}
    \caption{Clean execution snapshots on the \textbf{LIBERO-Long} suite. Each of the 10 rows represents a distinct manipulation task, displaying 9 equidistant frames from task initialization to completion.}
    \label{fig:clean_spatial}
\end{figure*}

\begin{figure*}[p]
    \centering
    \includegraphics[page=2, width=\textwidth]{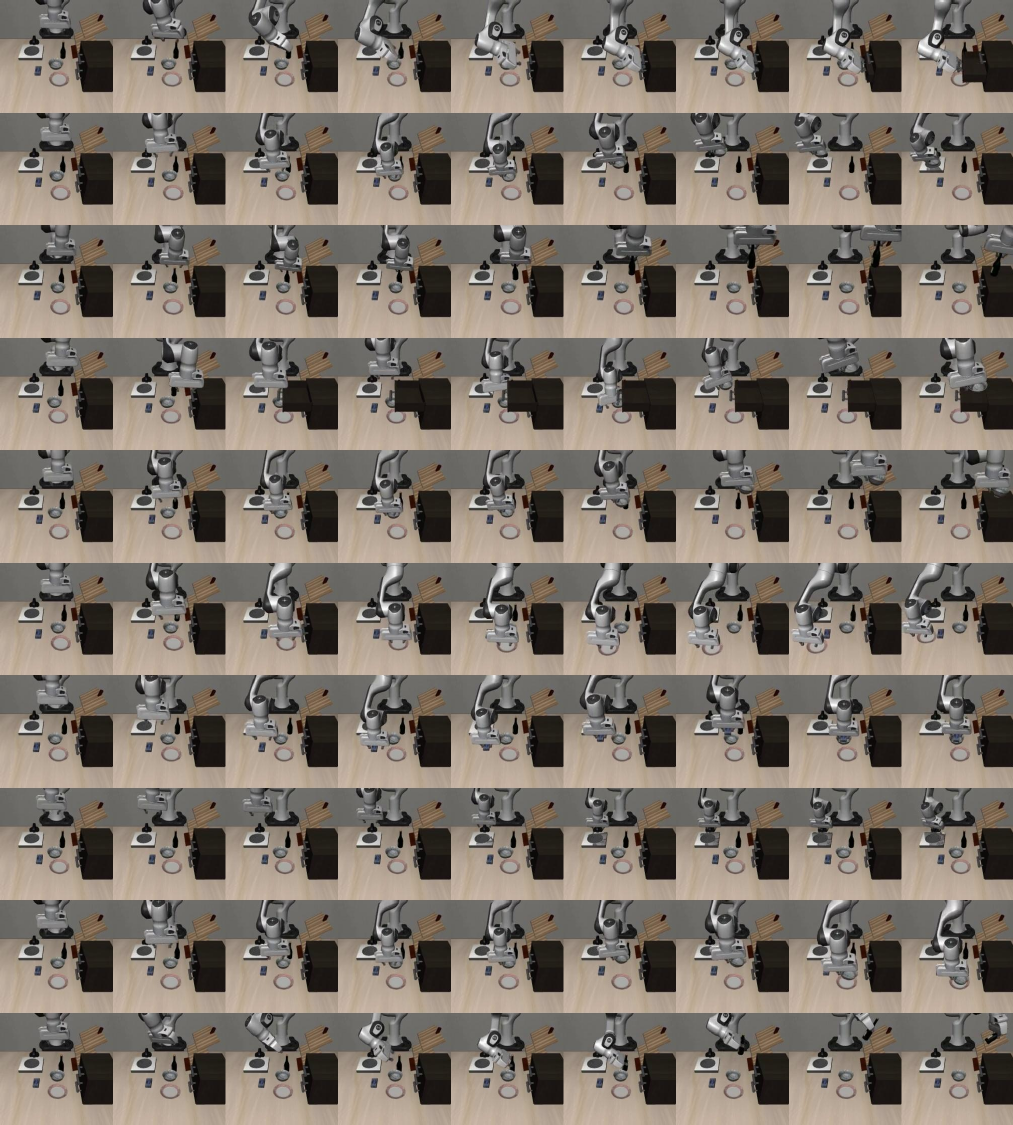}
    \caption{Clean execution snapshots on the \textbf{LIBERO-Goal} suite. Each of the 10 rows represents a distinct manipulation task, displaying 9 equidistant frames from task initialization to completion.}
    \label{fig:clean_object}
\end{figure*}

\begin{figure*}[p]
    \centering
    \includegraphics[page=3, width=\textwidth]{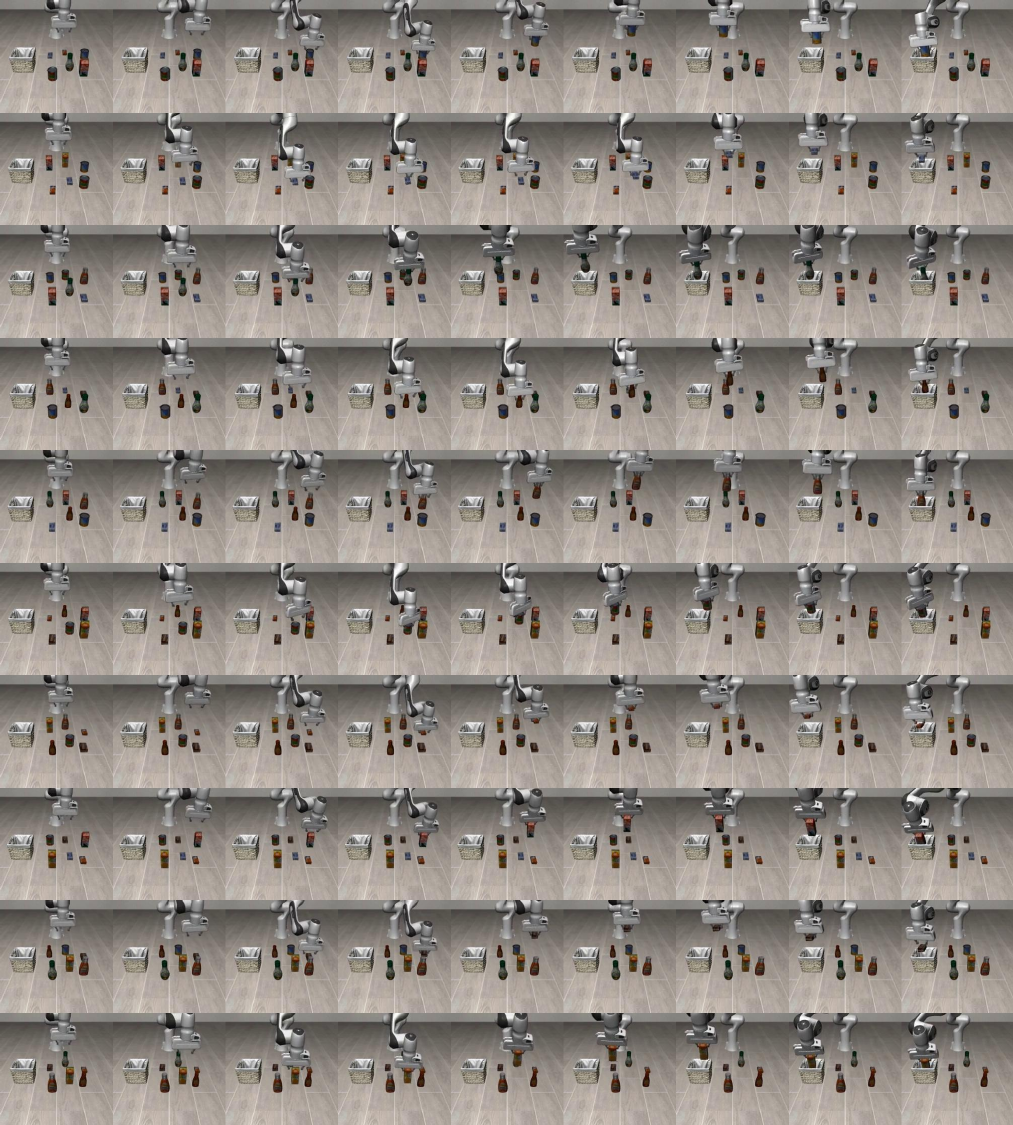}
    \caption{Clean execution snapshots on the \textbf{LIBERO-Object} suite. Each of the 10 rows represents a distinct manipulation task, displaying 9 equidistant frames from task initialization to completion.}
    \label{fig:clean_goal}
\end{figure*}

\begin{figure*}[p]
    \centering
    \includegraphics[page=4, width=\textwidth]{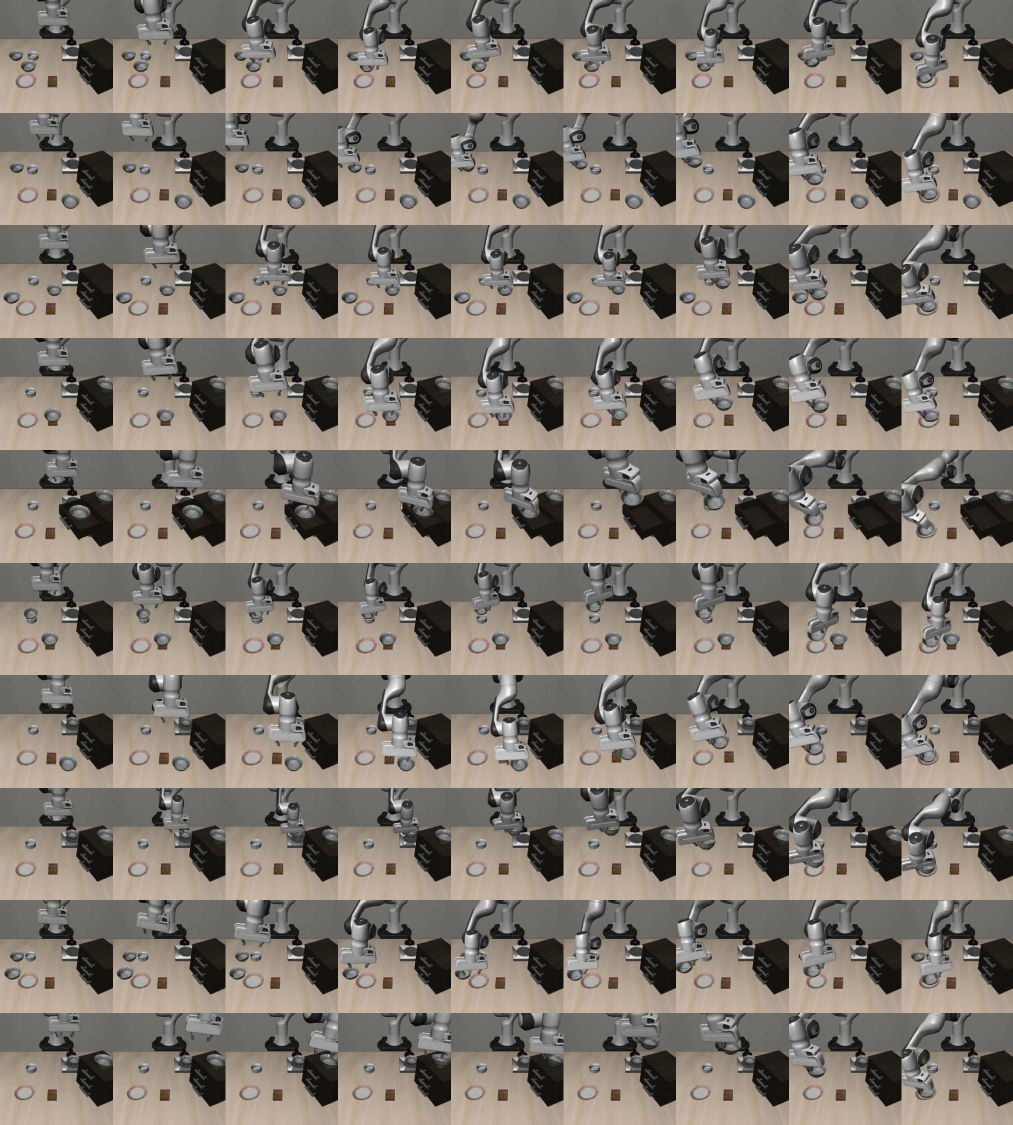}
    \caption{Clean execution snapshots on the \textbf{LIBERO-Spatial} suite. Each of the 10 rows represents a distinct manipulation task, displaying 9 equidistant frames from task initialization to completion.}
    \label{fig:clean_long}
\end{figure*}

\begin{figure*}[p]
    \centering
    \includegraphics[page=5, width=\textwidth]{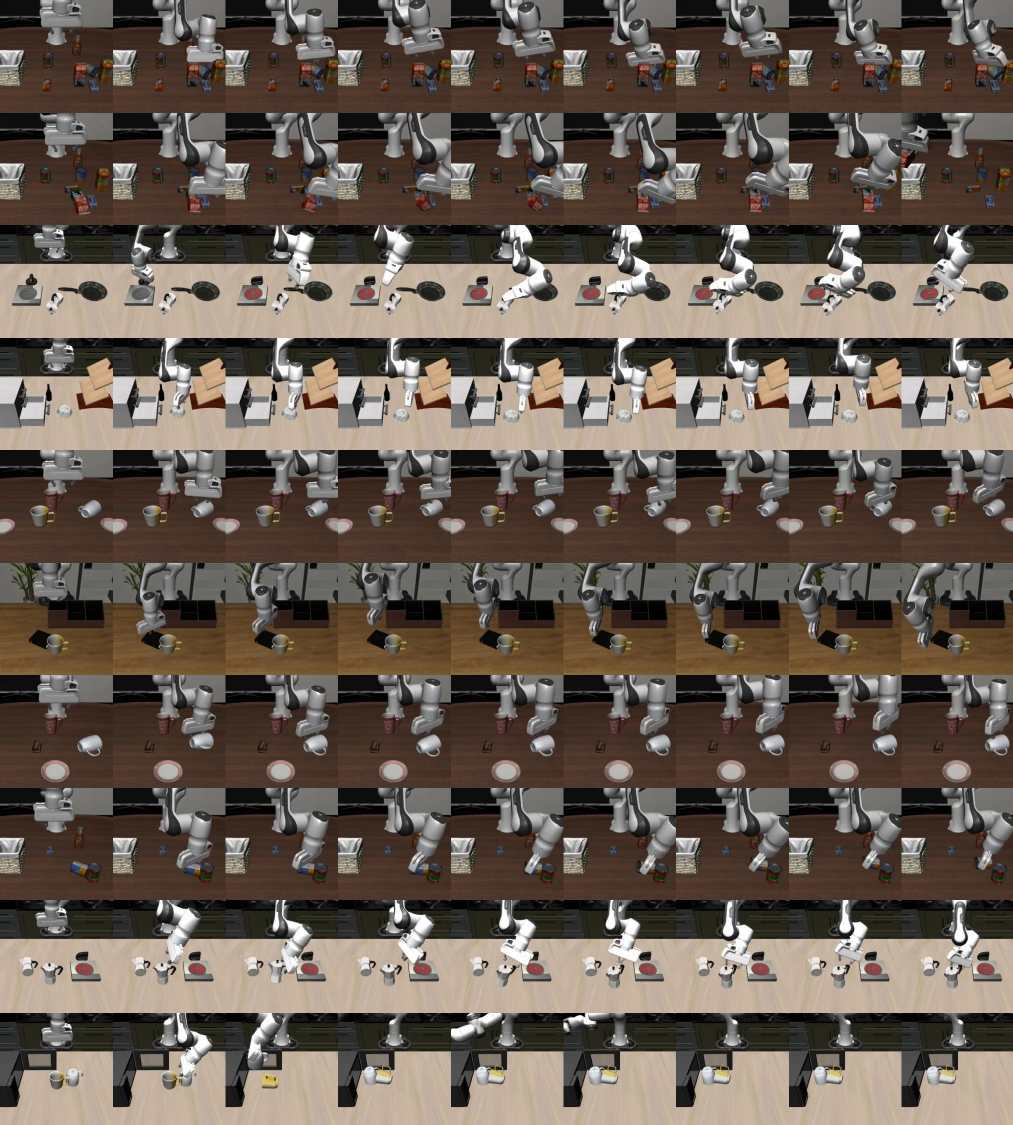}
    \caption{Failure modes under \textbf{object 3D transformations} on the \textbf{LIBERO-10} suite. The induced spatial misalignment consistently leads to grasp misses.}
    \label{fig:pose_spatial}
\end{figure*}

\begin{figure*}[p]
    \centering
    \includegraphics[page=6, width=\textwidth]{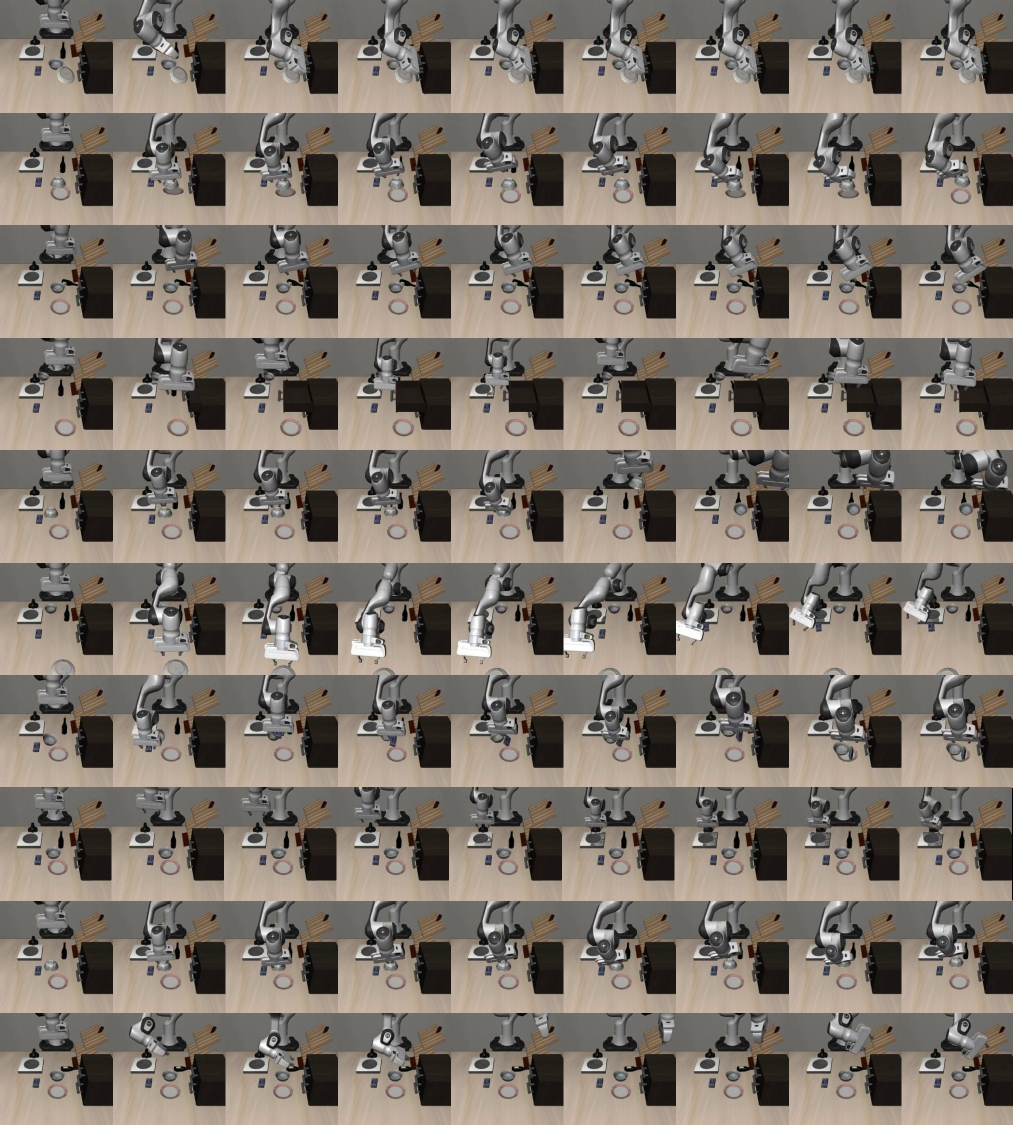}
    \caption{Failure modes under \textbf{object 3D transformations} on the \textbf{LIBERO-Goal} suite. The induced spatial misalignment consistently leads to grasp misses.}
    \label{fig:pose_object}
\end{figure*}

\begin{figure*}[p]
    \centering
    \includegraphics[page=7, width=\textwidth]{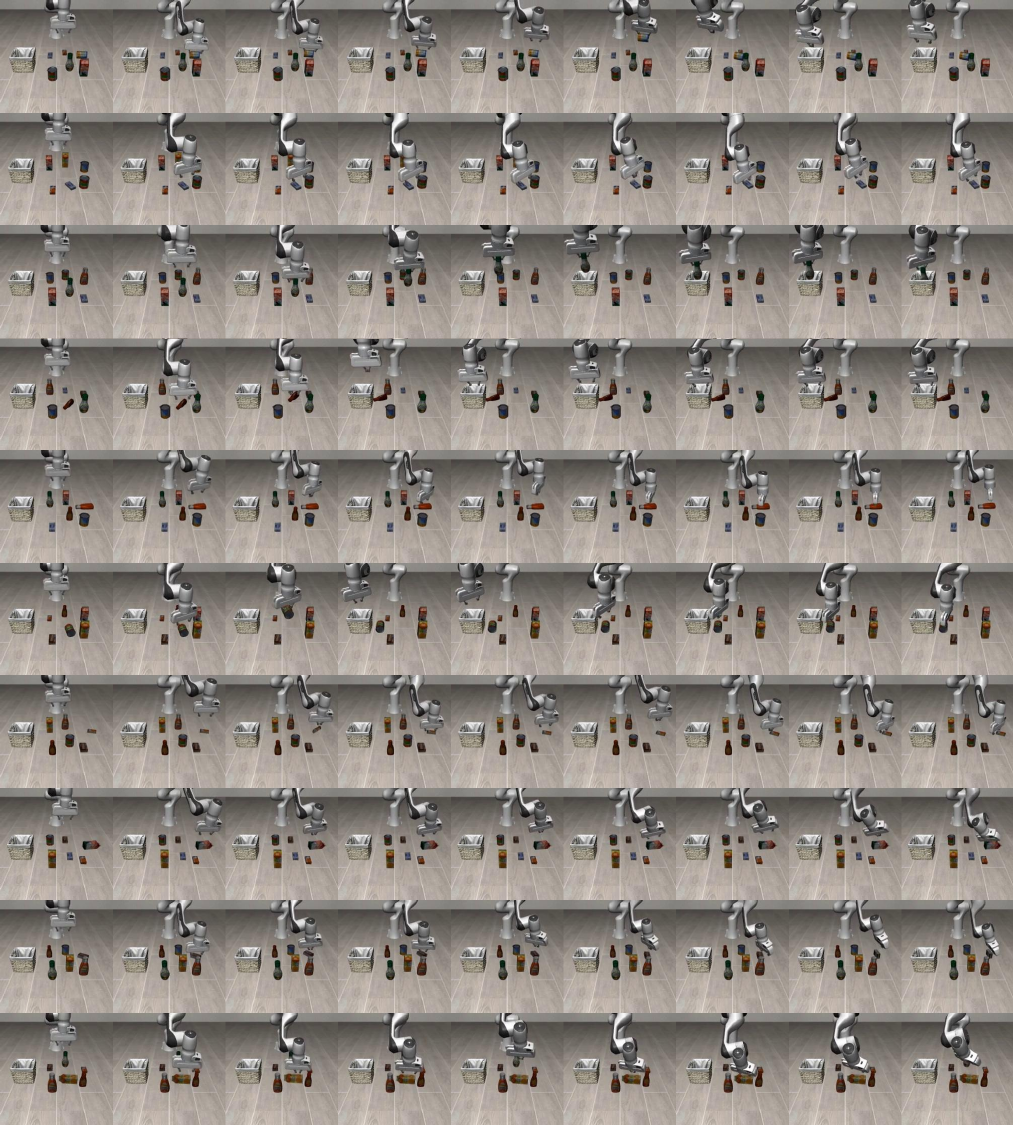}
    \caption{Failure modes under \textbf{object 3D transformations} on the \textbf{LIBERO-Object} suite. The induced spatial misalignment consistently leads to grasp misses.}
    \label{fig:pose_goal}
\end{figure*}

\begin{figure*}[p]
    \centering
    \includegraphics[page=8, width=\textwidth]{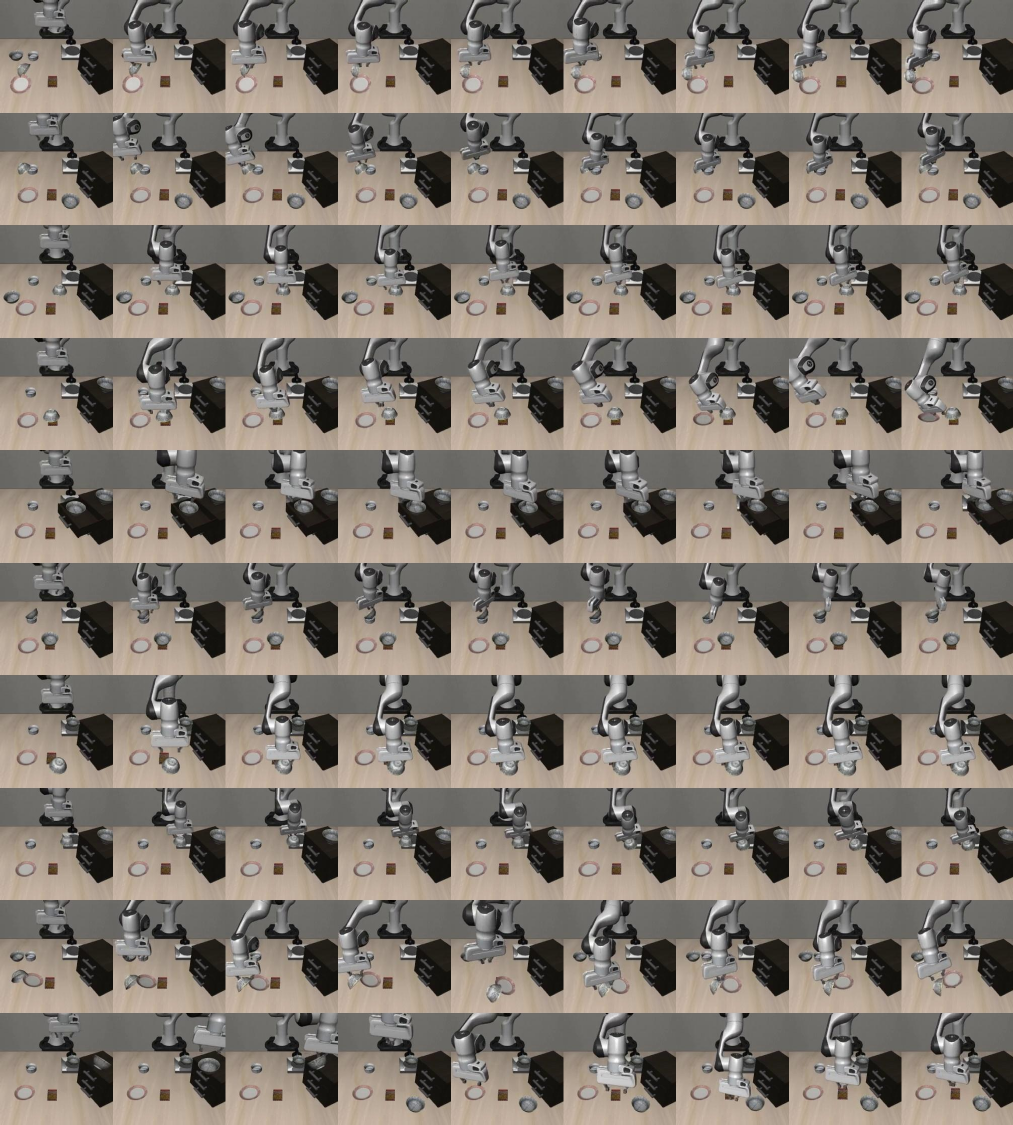}
    \caption{Failure modes under \textbf{object 3D transformations} on the \textbf{LIBERO-Spatial} suite. The induced spatial misalignment consistently leads to grasp misses.}
    \label{fig:pose_long}
\end{figure*}

\begin{figure*}[p]
    \centering
    \includegraphics[page=9, width=\textwidth]{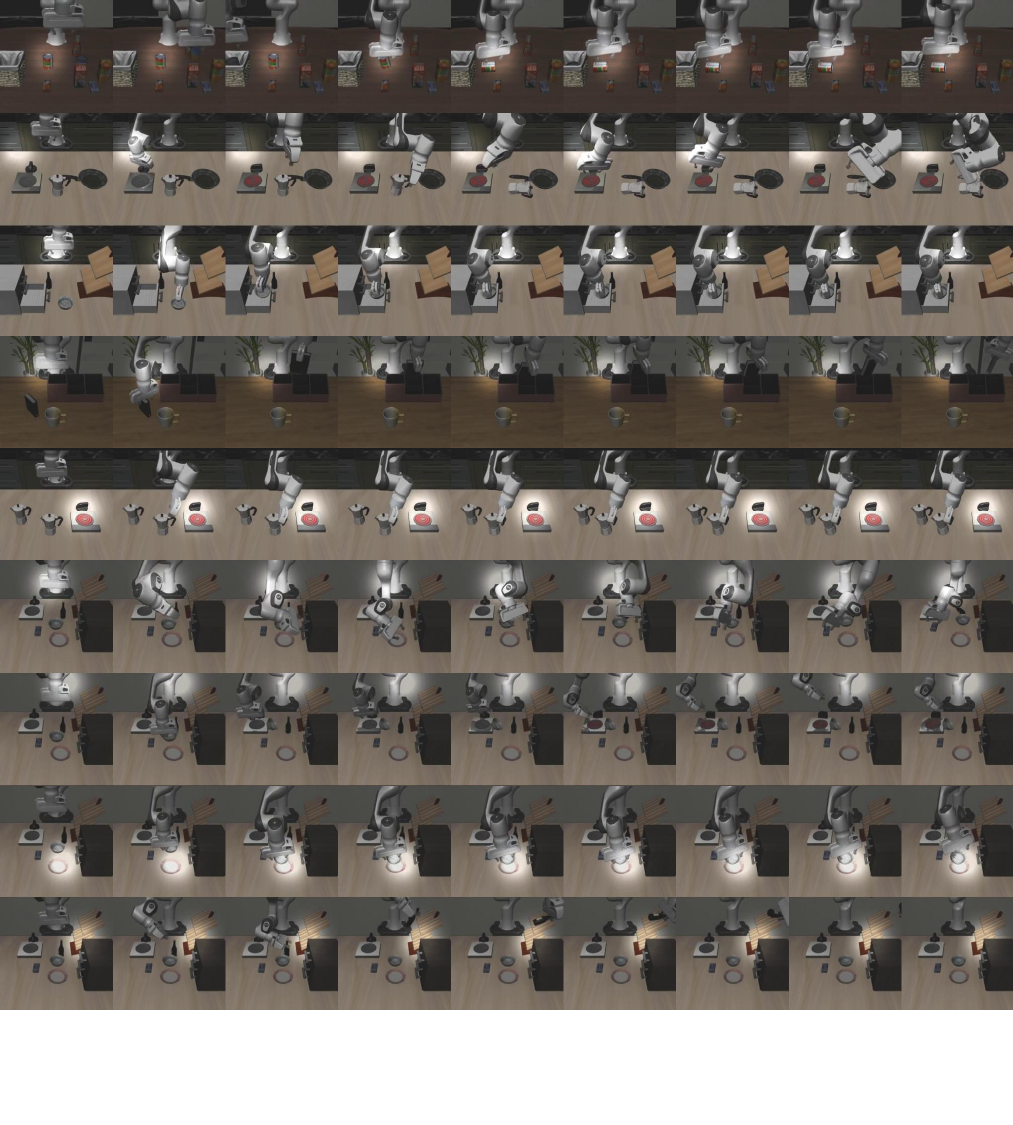}
    \caption{Failure modes under \textbf{illumination variations} (Illumination Variation 1). This visualization aggregates representative compromised tasks across different LIBERO suites, showing how obscured visual perception leads to premature task termination.}
    \label{fig:illum_var_1}

\end{figure*}

\begin{figure*}[p]
    \centering
    \includegraphics[page=10, width=\textwidth]{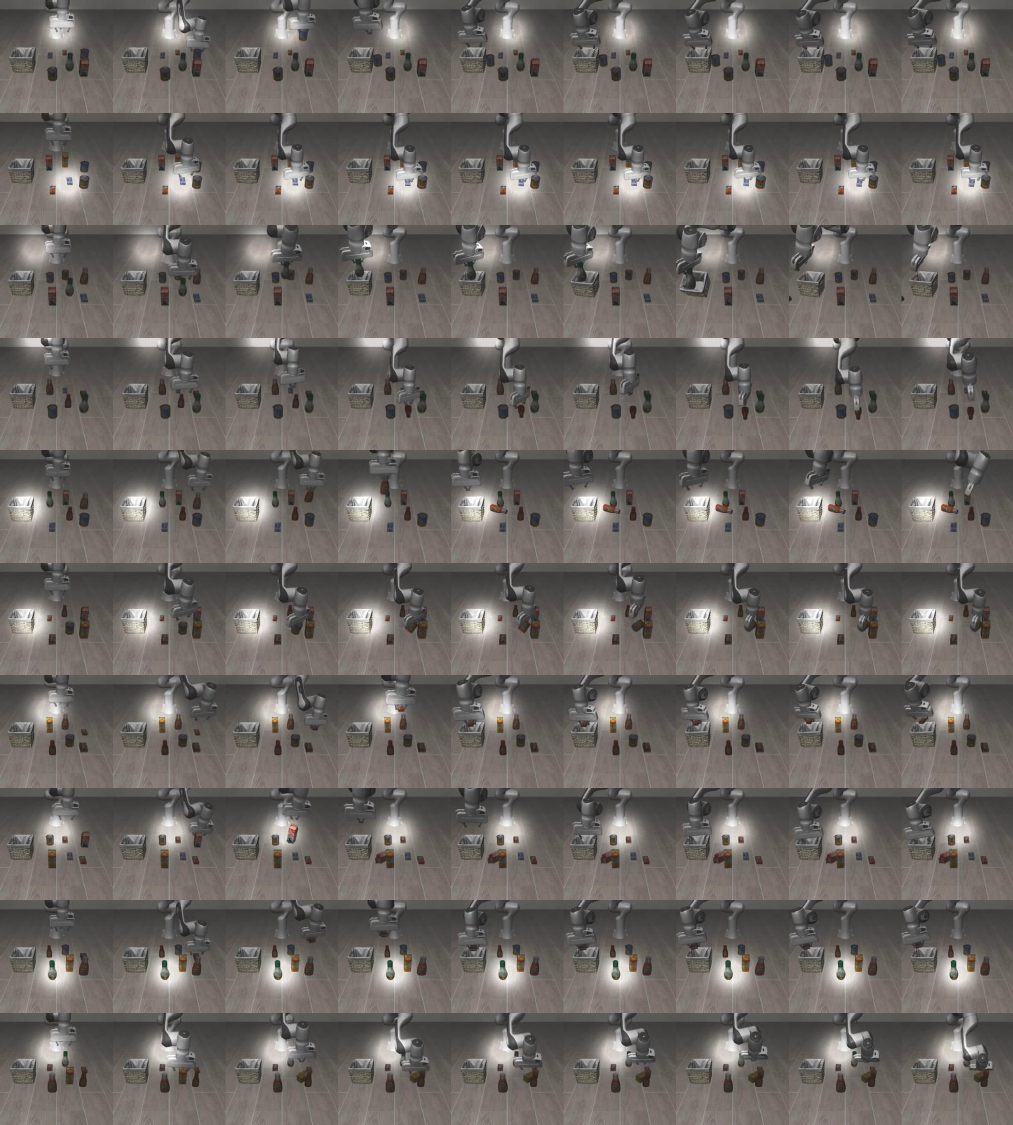}
    \caption{Failure modes under \textbf{illumination variations} (Illumination Variation 2). This visualization aggregates representative compromised tasks across different LIBERO suites, showing how obscured visual perception leads to premature task termination.}
    \label{fig:illum_var_2}
\end{figure*}

\begin{figure*}[p]
    \centering
    \includegraphics[page=11, width=\textwidth]{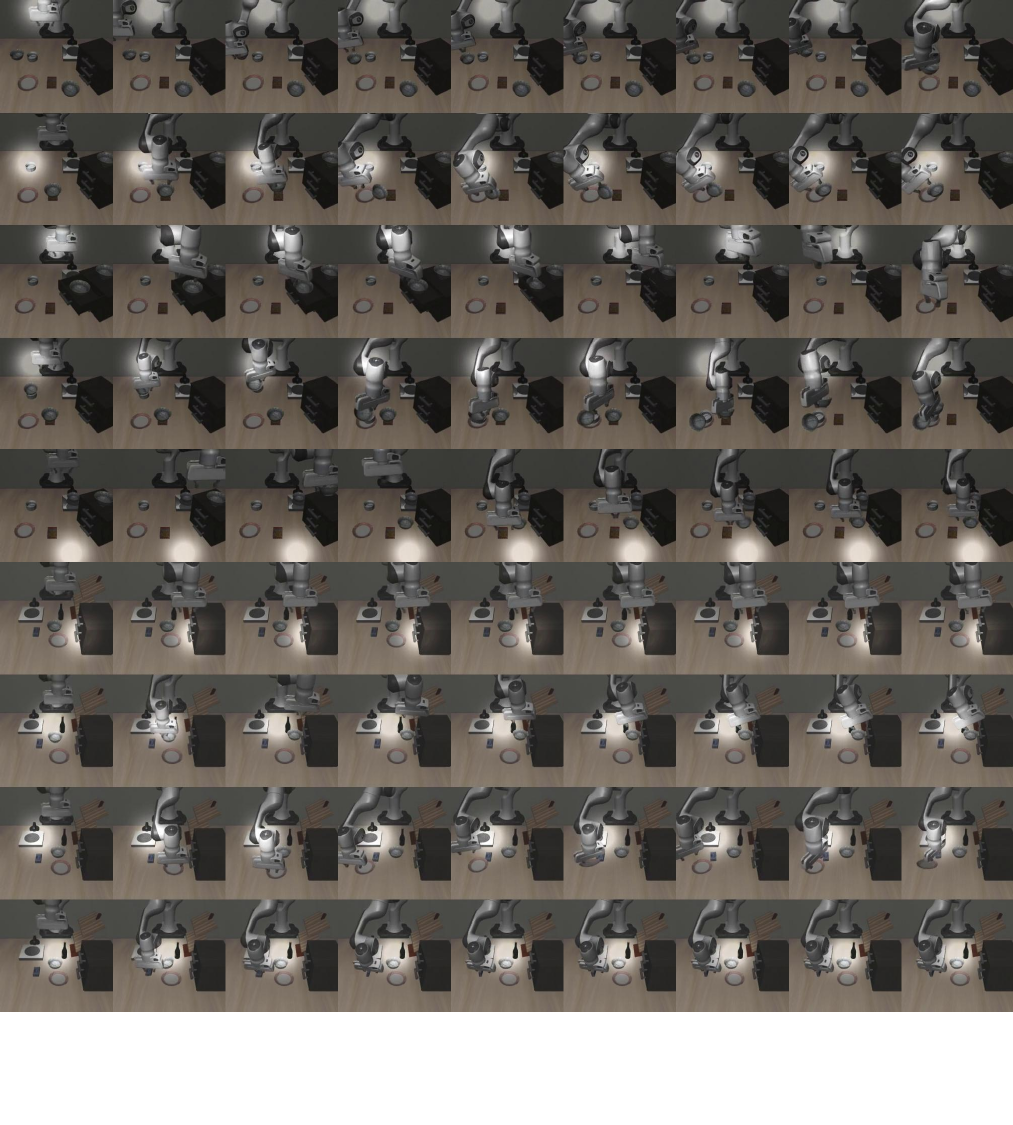}
    \caption{Failure modes under \textbf{illumination variations} (Illumination Variation 3). This visualization aggregates representative compromised tasks across different LIBERO suites, showing how obscured visual perception leads to premature task termination.}
    \label{fig:illum_var_3}
\end{figure*}

\begin{figure*}[p]
    \centering
    \includegraphics[page=12, width=\textwidth]{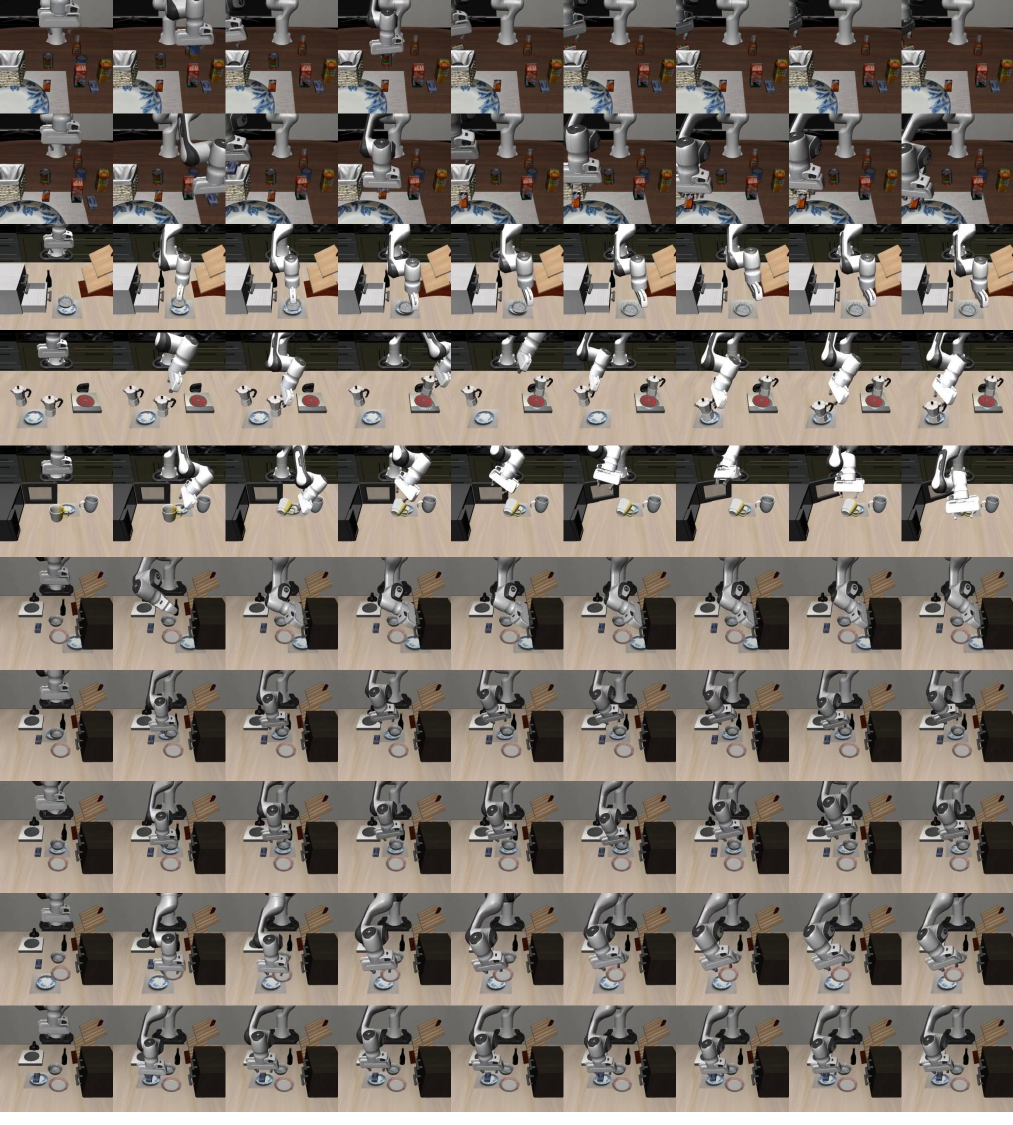}
    \caption{Failure modes under \textbf{adversarial patches} (Patch Variation 1). This visualization aggregates compromised tasks across different LIBERO suites, demonstrating how the adversarial patch distorts the model's perception of the placement region, misleading the robot to incorrectly place the target object directly onto the patch rather than the intended goal location.}
    \label{fig:patch_var_1}
\end{figure*}

\begin{figure*}[p]
    \centering
    \includegraphics[page=13, width=\textwidth]{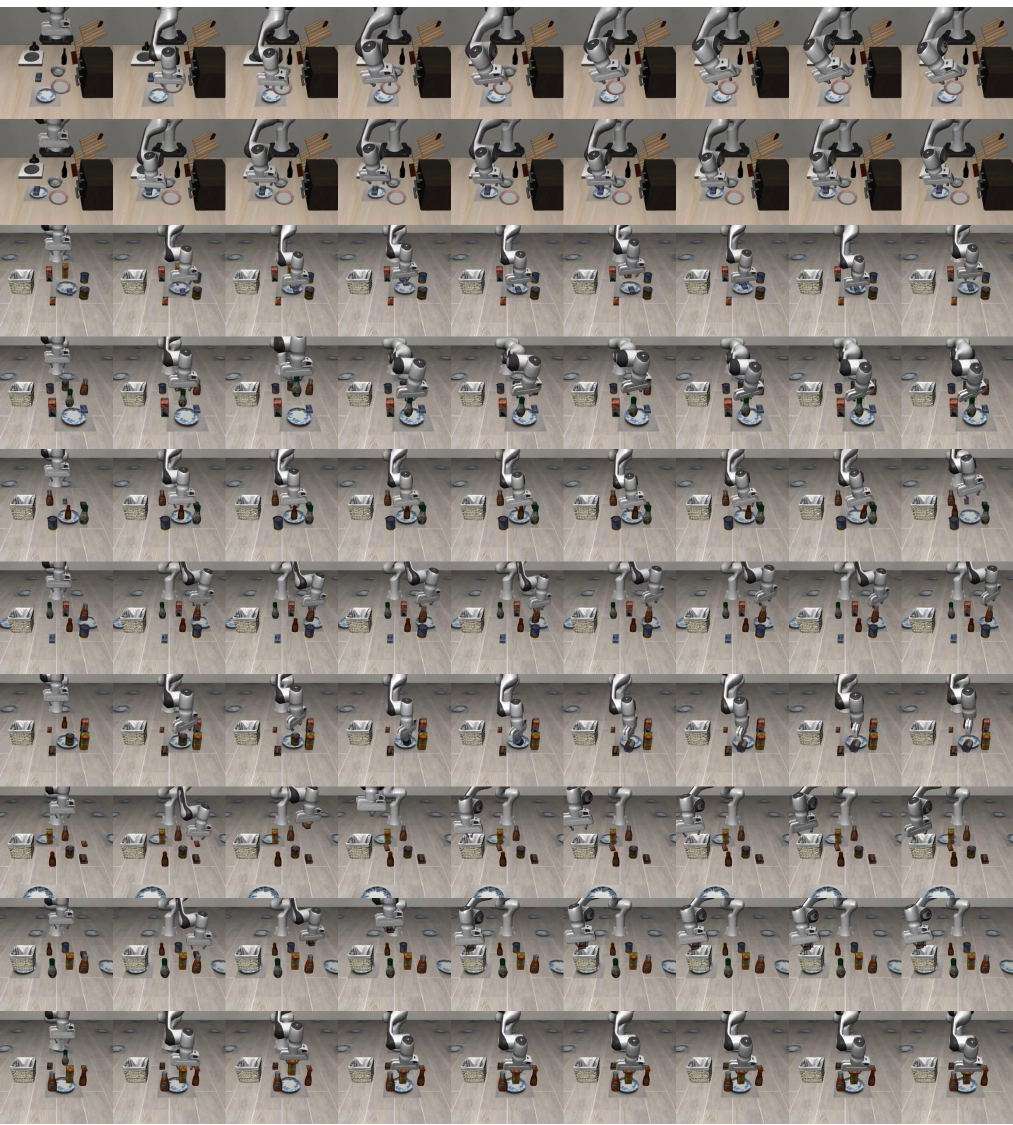}
    \caption{Failure modes under \textbf{adversarial patches} (Patch Variation 2). This visualization aggregates compromised tasks across different LIBERO suites, demonstrating how the adversarial patch distorts the model's perception of the placement region, misleading the robot to incorrectly place the target object directly onto the patch rather than the intended goal location.}
    \label{fig:patch_var_2}
\end{figure*}

\begin{figure*}[p]
    \centering
    \includegraphics[page=14, width=\textwidth]{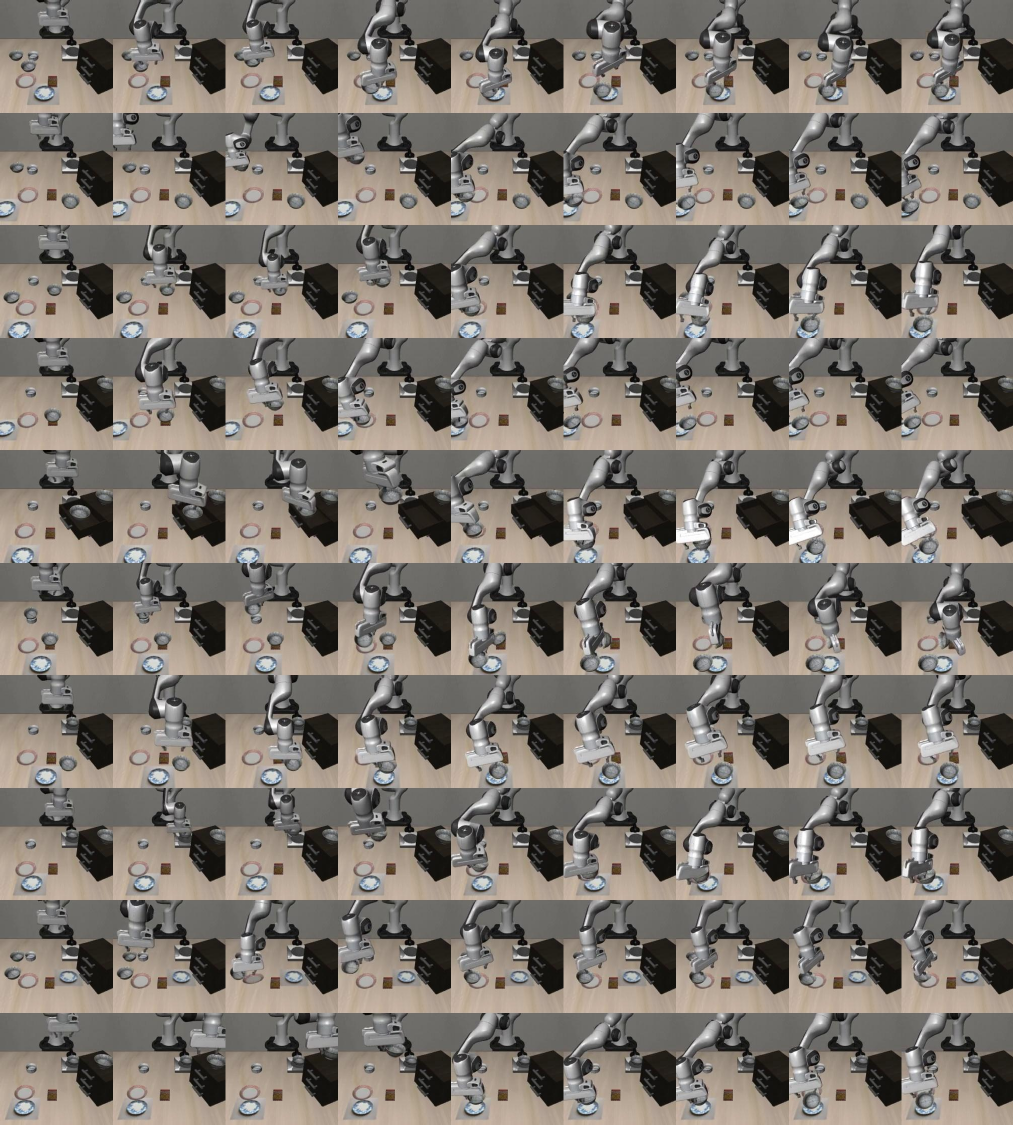}
    \caption{Failure modes under \textbf{adversarial patches} (Patch Variation 3). This visualization aggregates compromised tasks across different LIBERO suites, demonstrating how the adversarial patch distorts the model's perception of the placement region, misleading the robot to incorrectly place the target object directly onto the patch rather than the intended goal location.}
    \label{fig:patch_var_3}
\end{figure*}

For each task scenario, we visually compare the $\pi_{0.5}$ robot's execution under the clean (unperturbed) environment (Fig.~\ref{fig:clean_spatial} -- \ref{fig:clean_long}) against the failure cases compromised by our generated worst-case variations. Specifically, the following execution snapshots demonstrate how \textbf{object 3D transformations} (Fig.~\ref{fig:pose_spatial} -- \ref{fig:pose_long}) consistently induce spatial misalignment and grasp misses, how \textbf{illumination variations} (Fig.~\ref{fig:illum_var_1} -- \ref{fig:illum_var_3}) obscure the model's visual perception leading to premature task termination, and how \textbf{adversarial patches} (Fig.~\ref{fig:patch_var_1} -- \ref{fig:patch_var_3}) distort the perception of the placement region, misleading the robot to deposit the target object directly onto the patch rather than the intended goal location. 

Please note that for illumination and adversarial patch variations, due to the selective vulnerability of certain tasks to specific physical attacks, we aggregate successful attack scenarios across different LIBERO suites into composite visualizations. These extended visual results corroborate our quantitative findings, providing a comprehensive and intuitive view of the systemic fragilities inherent in current VLA architectures, such as the $\pi_{0.5}$ model, when deployed in physically variable environments.

\section*{Appendix C. Additional Adversarial Training Results on the LIBERO Benchmark}

To demonstrate that the physical variations discovered by Eva-VLA can serve as constructive feedback for model refinement, we conduct targeted adversarial training. This section details the data collection protocol and the fine-tuning hyperparameter configurations used to enhance the robustness of the $\pi_{0.5}$ model.

\subsection*{C.1. Adversarial Data Collection Protocol}

A critical challenge in adversarial training for robotic manipulation is acquiring high-quality demonstration trajectories under perturbed conditions. Depending on the nature of the physical variation, we employ two distinct data collection strategies to ensure task success while maintaining a balanced dataset:

\vspace{1em}
\noindent \textbf{Non-Spatial Variations (Illumination and Adversarial Patches):} 
Variations such as lighting changes ($\boldsymbol{\Lambda}$) and the introduction of adversarial patches ($\boldsymbol{\phi}$) alter the visual observation space but do not modify the underlying physical geometry or the required kinematics of the target object. Consequently, the original manipulation trajectories remain physically valid. For these variations, we efficiently synthesize paired adversarial examples by initializing the environment with the compromised visual parameters and directly applying the successful expert trajectories previously collected in the clean environments.

\vspace{1em}
\noindent \textbf{Spatial Variations (Object 3D Transformations):} 
Conversely, altering the 6-DoF pose of the target object ($\boldsymbol{\Theta}$) fundamentally changes the spatial requirements of the task. Replaying trajectories from the clean environment in these altered states inevitably leads to grasp misses or task failures. To address this, we introduce manual human-in-the-loop teleoperation via keyboard input to collect new, successful trajectories in the spatially perturbed environments. 

To ensure a fair and balanced comparison across all attack modalities, we strictly collect exactly 20 successful trajectories for each specific task under every variation type.

\subsection*{C.2. Fine-Tuning Strategy and Hyperparameters}

To prevent catastrophic forgetting and maintain the model's proficiency on standard, unperturbed tasks (as evidenced by the marginal clean failure rate increase from 4.0\% to 5.0\% discussed in the main text), we adopt a data-mixing strategy. The newly collected adversarial trajectories are uniformly mixed with the original clean LIBERO dataset during the fine-tuning phase.

The fine-tuning process initializes with the pre-trained $\pi_{0.5}$ base weights. The model is trained for 30,000 steps using an AdamW optimizer with a learning rate of $5 \times 10^{-5}$ and a 10,000-step linear warmup. The detailed training configurations are summarized in Table~\ref{tab:hyperparameters}.

\begin{table}[t]
\centering
\caption{Hyperparameters for Adversarial Fine-Tuning of the $\pi_{0.5}$ Model.}
\label{tab:hyperparameters}
\begin{tabular}{lc}
\toprule
\textbf{Hyperparameter} & \textbf{Value} \\
\midrule
Base Model Architecture & $\pi_{0.5}$ \\
Action Horizon & 10 \\
Prompting Strategy & Prompt from task \\
Batch Size & 256 \\
Optimizer & AdamW \\
Gradient Clipping Norm & 1.0 \\
Peak Learning Rate & $5 \times 10^{-5}$ \\
Learning Rate Schedule & Cosine Decay (flat post-warmup) \\
Warmup Steps & 10,000 \\
Total Training Steps & 30,000 \\
EMA Decay & 0.999 \\
\bottomrule
\end{tabular}
\end{table}

\section*{Appendix D. Additional Real-World Results}

To complement the findings presented in the main text, this appendix provides a detailed quantitative breakdown and extended visual comparisons of our real-world physical variation experiments.

\subsection*{D.1. Real-World Experimental Setup}
We systematically evaluate the real-world robustness of Vision-Language-Action (VLA) architectures by focusing on the representative ``\textit{put A into B}'' instruction paradigm. For these experiments, we deploy the OpenVLA\cite{kim2024openvla} model. 

To establish a robust baseline and adapt the model to our specific workspace, we define three distinct manipulation tasks, each corresponding to one of the targeted physical variations. For each task, we first collect 50 expert trajectories in a clean, unperturbed environment to serve as the fine-tuning dataset. Following the fine-tuning phase, we conduct rigorous evaluations consisting of 20 independent execution trials per task under both the clean baseline condition and the respective physically perturbed condition.

\subsection*{D.2. Quantitative Evaluation}
The quantitative results of our real-world evaluations are summarized in Table~\ref{tab:real_world_results}. The data explicitly demonstrates the severe vulnerabilities of the fine-tuned OpenVLA model when exposed to our proposed physical variations:

\begin{itemize}
    \item \textbf{Put the toy cat into the plate (Object 3D Transformations):} Altering the spatial pose of the target object completely paralyzed the model's execution capabilities, plummeting the task success rate from a clean baseline of 45\% to 0\%.
    \item \textbf{Put the peppers into the plate (Illumination Variations):} Extreme lighting changes severely obscured the model's visual perception, resulting in a dramatic performance drop from 50\% success in the clean environment to just 10\%.
    \item \textbf{Put the banana into the plate (Adversarial Patches):} The introduction of adversarial patches into the scene effectively hijacked the model's scene understanding, reducing the success rate from 50\% to a mere 5\%.
\end{itemize}

\begin{table}[ht]
\centering
\renewcommand{\arraystretch}{1.4} % 稍微再增加一点行高，防止换行的文字上下太拥挤
\caption{Real-world evaluation success rates of the fine-tuned OpenVLA model across 20 trials per task. Each task targets a specific physical variation under the ``\textit{put A into B}'' paradigm.}
\label{tab:real_world_results}
\begin{tabular}[width=\linewidth]{cccc}
\toprule
\textbf{Task Instruction} & \textbf{Variation Type} & \textbf{Clean} & \textbf{Perturbed} \\
\midrule
Put the toy cat into the plate & 3D Trans. ($\boldsymbol{\Theta}$) & 45\% (9/20) & \textbf{0\% (0/20)} \\
Put the peppers into the plate & Illum. ($\boldsymbol{\Lambda}$) & 50\% (10/20) & \textbf{10\% (2/20)} \\
Put the banana into the plate  & Adv.Patch ($\boldsymbol{\phi}$) & 50\% (10/20) & \textbf{5\% (1/20)} \\
\bottomrule
\end{tabular}
\end{table}

\subsection*{D.3. Qualitative Visualizations}
As noted in the main text, the quantitative failures are frequently accompanied by erratic physical behaviors. Under these perturbed conditions, the OpenVLA model consistently exhibits unstable, jerky, and oscillatory robotic motions, which pose significant risks in real-world deployments. 

The following figures (Fig.~\ref{fig:real_task1} - \ref{fig:real_task3}) provide visual comparisons using 36 uniformly sampled frames across the entire task duration. These composite snapshots contrast the smooth, successful execution in the clean environment against the corresponding failure modes induced by each physical variation.

% ==========================================
% 1. Clean Environments (Pages 1-4)
% ==========================================

% ==========================================
% 2. Pose Transformations (Pages 5-8)
% ==========================================

% ==========================================
% 3. Illumination Variations (Pages 9-11)
% ==========================================

% ==========================================
% 4. Adversarial Patches (Pages 12-14)
% ==========================================

% ==========================================
% Insert your real-world images here
% ==========================================

\begin{figure*}[htbp]
    \centering
    \includegraphics[width=\textwidth]{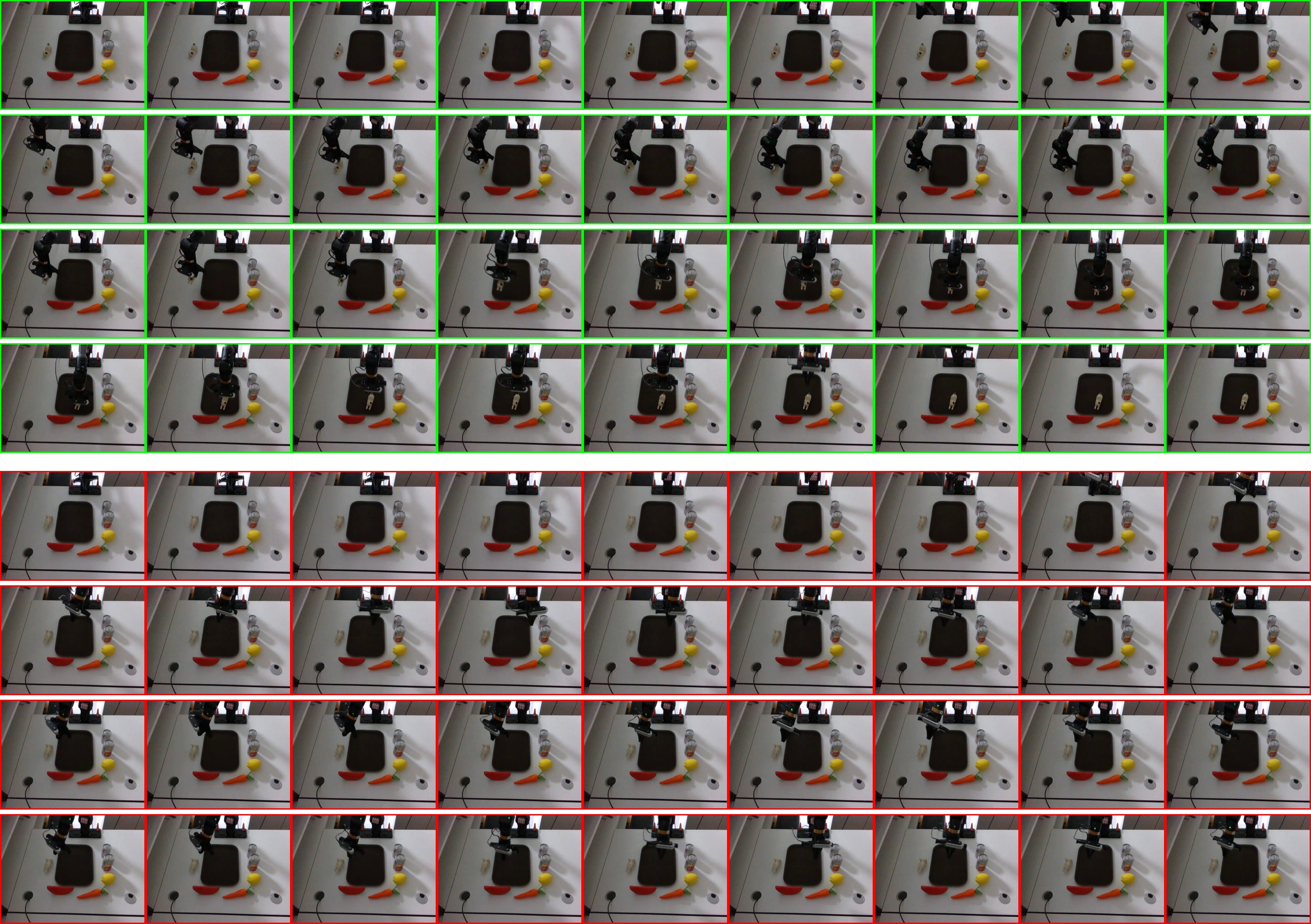}
    \caption{Real-world execution comparison for the ``\textit{put the toy cat into the plate}'' task. The top block (green borders) shows successful execution in the clean environment. The bottom block (red borders) demonstrates how \textbf{Object 3D Transformations} disrupt spatial manipulation, leading to a 0\% success rate.}
    \label{fig:real_task1}
\end{figure*}

\begin{figure*}[htbp]
    \centering
    \includegraphics[width=\textwidth]{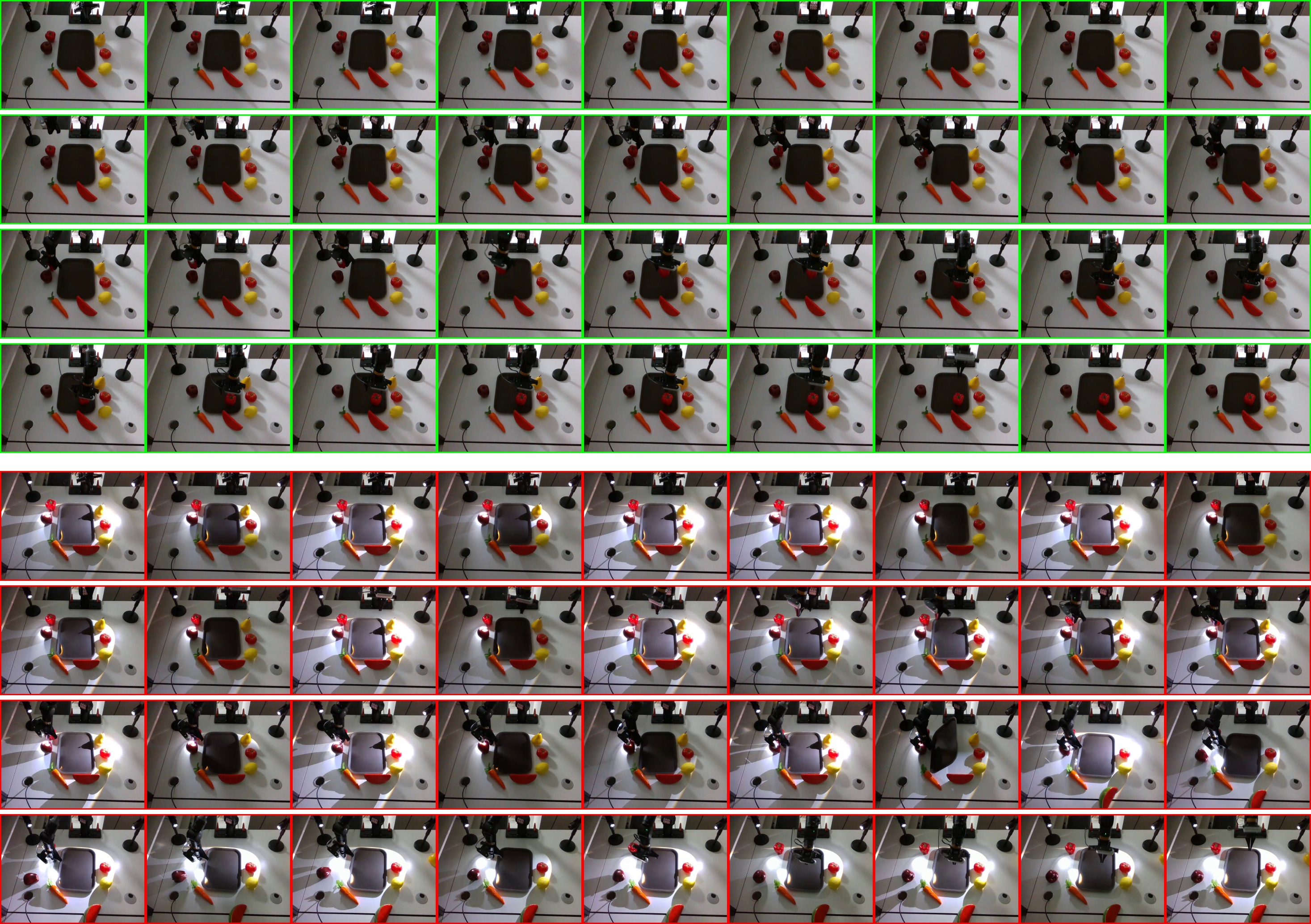}
    \caption{Real-world execution comparison for the ``\textit{put the peppers into the plate}'' task. The top block (green borders) shows successful execution in the clean environment. The bottom block (red borders) demonstrates how extreme \textbf{Illumination Variations} obscure visual perception.}
    \label{fig:real_task2}
\end{figure*}

\begin{figure*}[htbp]
    \centering
    \includegraphics[width=\textwidth]{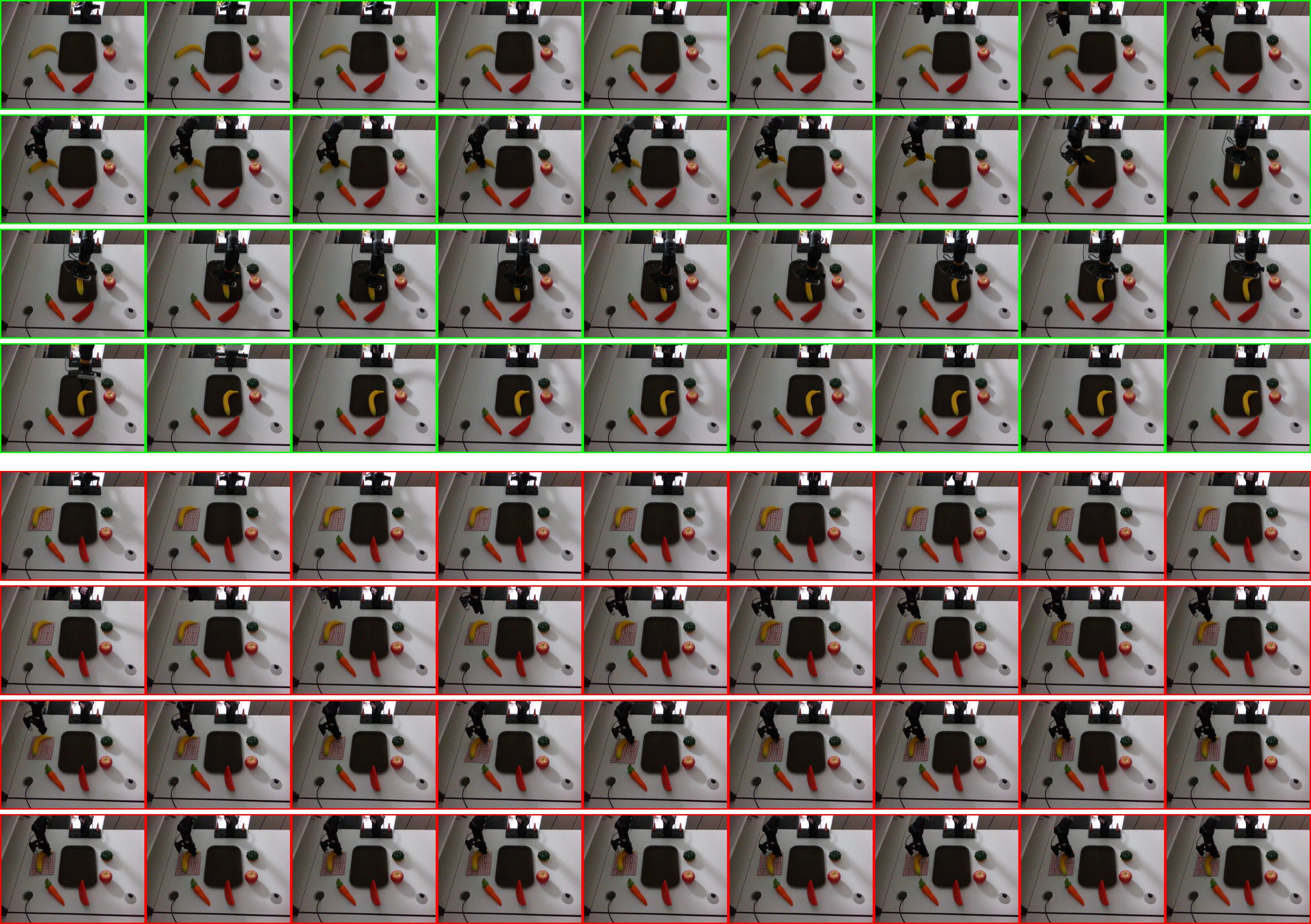}
    \caption{Real-world execution comparison for the ``\textit{put the banana into the plate}'' task. The top block (green borders) shows successful execution in the clean environment. The bottom block (red borders) demonstrates how \textbf{Adversarial Patches} severely interfere with scene understanding and placement intent.}
    \label{fig:real_task3}
\end{figure*}

\clearpage
\bibliographystyle{splncs04}
\bibliography{main}